\documentclass[10pt,twocolumn,letterpaper]{article}

\usepackage{CVPR/cvpr}              

\usepackage{abstract}
\usepackage{amsfonts}       
\usepackage{amsmath}
\usepackage{amssymb}
\usepackage{array}
\usepackage[accsupp]{axessibility} 
\usepackage{blkarray}
\usepackage{bm}
\usepackage{booktabs}       
\usepackage{capt-of}
\usepackage{color}
\usepackage{comment}
\usepackage{epsfig}
\usepackage[export]{adjustbox}
\usepackage{float}
\usepackage[T1]{fontenc}    
\usepackage{graphicx}
\usepackage[utf8]{inputenc} 
\usepackage{microtype}      
\usepackage{multirow}
\usepackage{nicefrac}       
\usepackage{numprint}
\usepackage{pifont}
\usepackage{subcaption}
\usepackage{subfiles}
\usepackage{tabularx}
\usepackage{tikz}
\usepackage{times}
\usepackage{url}            
\usepackage{xcolor}         
\usepackage{xspace}
\usepackage{wrapfig}
\usepackage[normalem]{ulem}

\definecolor{cvprblue}{rgb}{0.21,0.49,0.74}
\usepackage[pagebackref,breaklinks,colorlinks,citecolor=cvprblue]{hyperref}

\newcolumntype{L}{>{\raggedright\arraybackslash}X}
\newcolumntype{C}{>{\centering\arraybackslash}X}
\newcolumntype{H}{>{\setbox0=\hbox\bgroup}c<{\egroup}@{}}

\usepackage[capitalize]{cleveref}
\crefname{section}{Sec.}{Secs.}
\Crefname{section}{Section}{Sections}
\Crefname{table}{Table}{Tables}
\crefname{table}{Tab.}{Tabs.}

\newcommand{\mycomment}[1]{}

\newcommand{\refsec}[1]{Sec.~\ref{#1}}
\newcommand{\reffig}[1]{Fig.~\ref{#1}}
\newcommand{\refeq}[1]{Eq.~\ref{#1}}
\newcommand{\reftab}[1]{Tab.~\ref{#1}}

\newcommand{\ba}{\begin{eqnarray*}}
\newcommand{\ea}{\end{eqnarray*}}

\newcommand{\DP}{DensePose~}

\newcommand{\sparsepose}{\textbf{S}}
\newcommand{\sparseposepix}{s} 
\newcommand{\sparseposechan}{S} 
\newcommand{\pixel}{\mathbf{x}}
\newcommand{\uv}{\mathbf{u}}

\def\paperID{14781} 
\def\confName{CVPR}
\def\confYear{2024}

\title{MeshPose: Unifying DensePose and 3D Body Mesh reconstruction}

\author{Eric-Tuan Lê\textsuperscript{1}\thanks{Equal contribution} \thanks{Work done while interning at Snap Inc.} $\;$
Antonis Kakolyris\textsuperscript{2}\footnotemark[1] $\;$
Petros Koutras\textsuperscript{2} $\;$
Himmy Tam\textsuperscript{2} $\;$
Efstratios Skordos\textsuperscript{2} \\
George Papandreou\textsuperscript{2} $\;$
R{\i}za Alp G\"uler\textsuperscript{2} $\;$
Iasonas Kokkinos\textsuperscript{2} \\
\vspace*{-0.8em} \\
\textsuperscript{1}University College London $\quad$ \textsuperscript{2}Snap Inc.
\\
{\href{https://meshpose.github.io/}{\small \url{meshpose.github.io}}}
}

\newcommand{\partpose}{VertexPose~}
\newcommand{\partposenogap}{VertexPose}

\begin{document}

\twocolumn[{
\renewcommand\twocolumn[1][]{#1}
\maketitle
\vspace{-1cm}
\begin{center}
    \centering
    \captionsetup{type=figure}    \includegraphics[width=\textwidth]{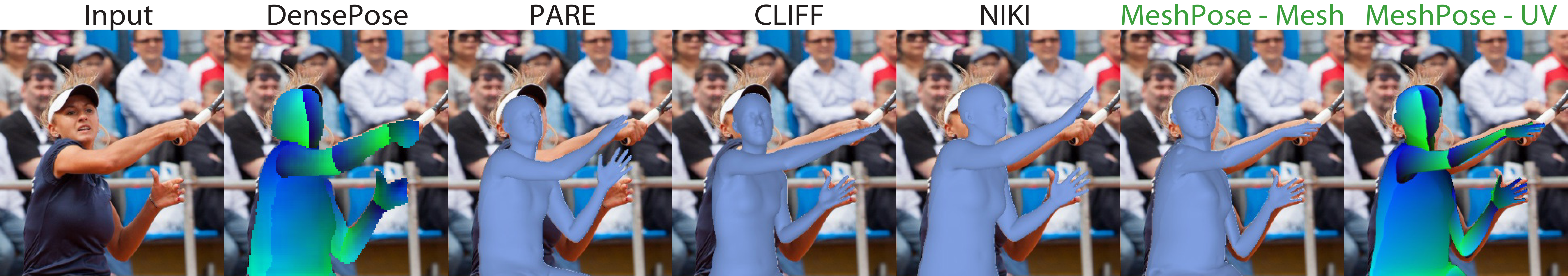}
    \captionof{figure}{DensePose prediction systems are pixel-accurate but do not provide a 3D mesh, while  human mesh recovery systems do not provide pixel-accurate 2D reprojection. We propose MeshPose, a novel human mesh recovery method that combines the benefits of both.}
\end{center}
}]
\thispagestyle{empty}
\saythanks

\begin{abstract}
DensePose provides a pixel-accurate association of images with 3D mesh coordinates, but does not provide a 3D mesh, while 
Human Mesh Reconstruction (HMR) systems have high 2D reprojection error,  as measured by \DP localization  metrics.
In this work we introduce MeshPose to jointly tackle DensePose and HMR. 
For this we first introduce new losses that allow us to use weak DensePose supervision to accurately localize in 2D a subset of the mesh vertices (`\partposenogap'). We then lift these vertices to 3D, yielding a low-poly body mesh (`MeshPose').  
Our system is trained in an end-to-end manner and is the first HMR method to attain competitive DensePose accuracy, while also being lightweight and amenable to efficient inference, making it suitable for real-time AR applications.
\end{abstract}

\newcommand{\reftable}[1]{Table~\ref{#1}}
\section{Introduction}

3D Human Mesh Reconstruction (HMR) has received increased attention thanks to its broad  AR/VR applications such as human-computer interaction, motion capture, entertainment/VFX and virtual try-on. 
Despite  rapid progress in HMR, mesh predictions with the current systems are still not pixel-accurate when projected back to the image domain. Mesh reconstruction evaluation is primarily 3D skeleton- or 3D mesh-based (measured in millimeters) and does not reflect 2D reprojection accuracy (in pixels). However, for persons close to the camera, small 3D errors result in visible 2D reprojection errors, for instance when users take selfie photos, as is the currently predominant use case for AR. If we want a 3D mesh that `looks good' when projected to 2D the present HMR method evaluation must be complemented by 2D reprojection metrics. 

Such errors are in the blindspot of 3D evaluation metrics and can break an AR experience such as virtual try-on. Errors in the 2D projection of a mesh are glaringly obvious (e.g. bags floating above the user's shoulders, coats that are too tight/too loose, misplacements around limbs etc.). DensePose has been a popular alternative to accurate warp tight garments to the user's body \cite{styleposegan, posewithstyle, du2024greatness}, however warping does not suffice for apparel (e.g. dress, coat, handbag) that protrudes from the user's body and requires proper 3D try-on.

\begin{figure}
\centering
\subfloat{{
 \includegraphics[width=0.23\textwidth]{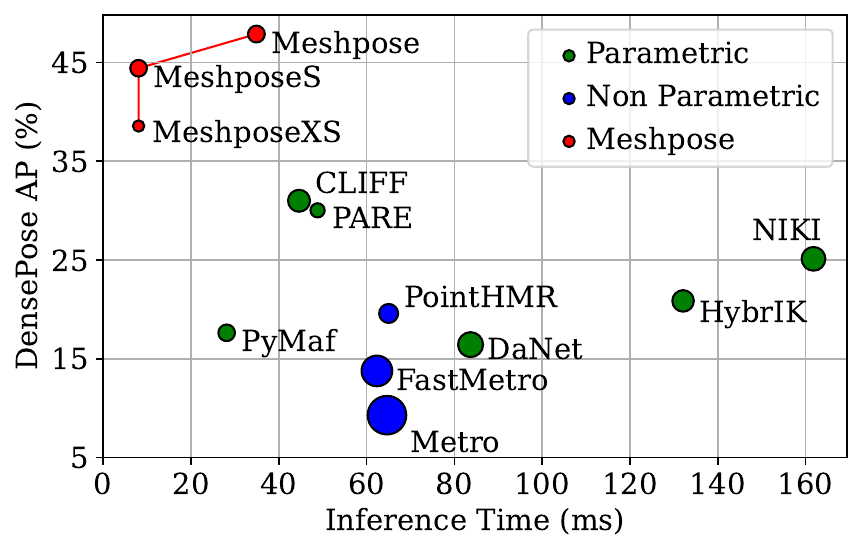}}} 
\subfloat{{ 
 \includegraphics[width=0.23\textwidth]{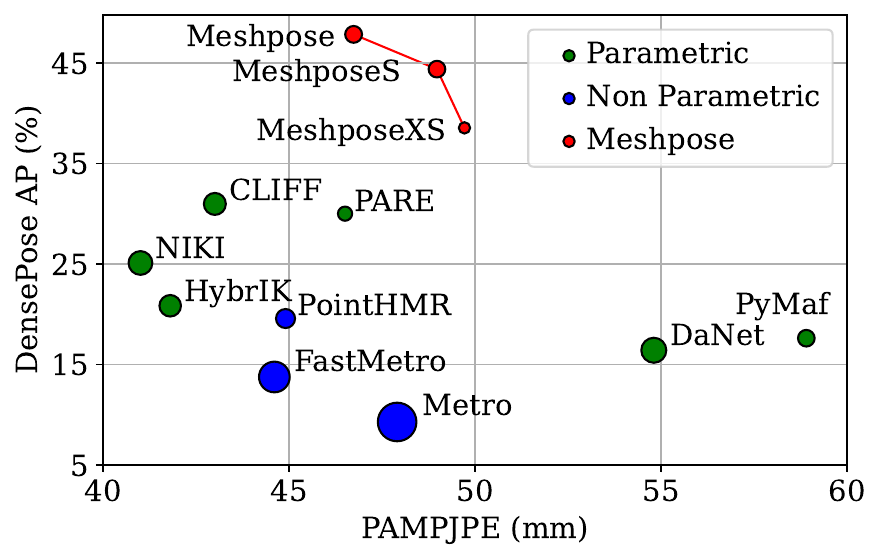}}}
    \caption{
    \textbf{Left}: Inference Time vs DensePose AP,
    \textbf{Right}: PA-MPJPE vs DensePose AP -- for both, top-left is best and radii are proportional to the sizes of the models (MB). Our approach outperforms HMR methods on DensePose metrics by more than 50\% while having close to state of the art 3D accuracy. By combining the highest FPS rate and small model size with state-of-art reprojection accuracy, our pipeline is well suited for mobile inference. 
    }
    \label{fig:scatter}
\end{figure}

Motivated by this observation, in this work we set out to bridge the gap between the DensePose and HMR systems. For this we revisit the DensePose task \cite{guler2018densepose} 
and show that we can use the \DP dataset to supervise a network that relies on a discrete, 3D vertex-based representation rather than relying on continuous UV prediction. The resulting system performs similarly to UV-based systems on the DensePose task while at the same time delivering an accurate 3D mesh. We call the resulting mesh prediction system ``MeshPose'' to indicate that it 
combines Mesh and DensePose prediction in a unified system. 

To achieve this we make the following contributions:
\begin{itemize} 
\item We introduce \partposenogap, a novel layer designed to 
predict the 2D projections of the vertices of a low poly 3D body mesh and simultaneously regress the per-pixel DensePose UV signal directly. This is accomplished through dense heatmaps that allow us to precisely pin down the pixel coordinates of a vertex. For this we introduce two new weakly supervised losses that rely on the mesh geometry to supervise \partpose through the DensePose dataset.
 
\item We then introduce MeshPose to form a 3D mesh out of the localized 2D vertices. This is accomplished by regressing per-vertex a depth, visibility and amodal 3D estimate. We lift visible vertices to 3D by concatenating the 2D position with their depth, and use the amodal estimate for invisible vertices.
\end{itemize}

Our results, shown in \reffig{fig:scatter} and in more detail in \reftable{tab:sota_3dpw} show that when assessed in terms of 2D \DP or even plain 2D pose estimation accuracy, other recent methods can use even 10$\times$ more parameters (e.g. MeshPoseXS vs Metro), or be 20$\times$ slower (e.g. MeshPoseXS vs NIKI) yet still result in substantially worse 2D reprojection metrics. At the same time our mesh reconstruction performance is comparable to most recent systems on 3DPW. Given the importance of 2D reprojection to the end-user experience in AR, we hope that our work will establish DensePose-based evaluation and training as a standard practice in future HMR works.  

Our video results, provided in the Supplement, complement these findings and indicate the temporal stability of our method even when applied frame-by-frame. 
Our method is lightweight, simple and directly amenable to real-time inference on mobile devices, making it a prime candidate for AR applications. 

\begin{figure*}
    \centering
    \includegraphics[width=\textwidth,clip]{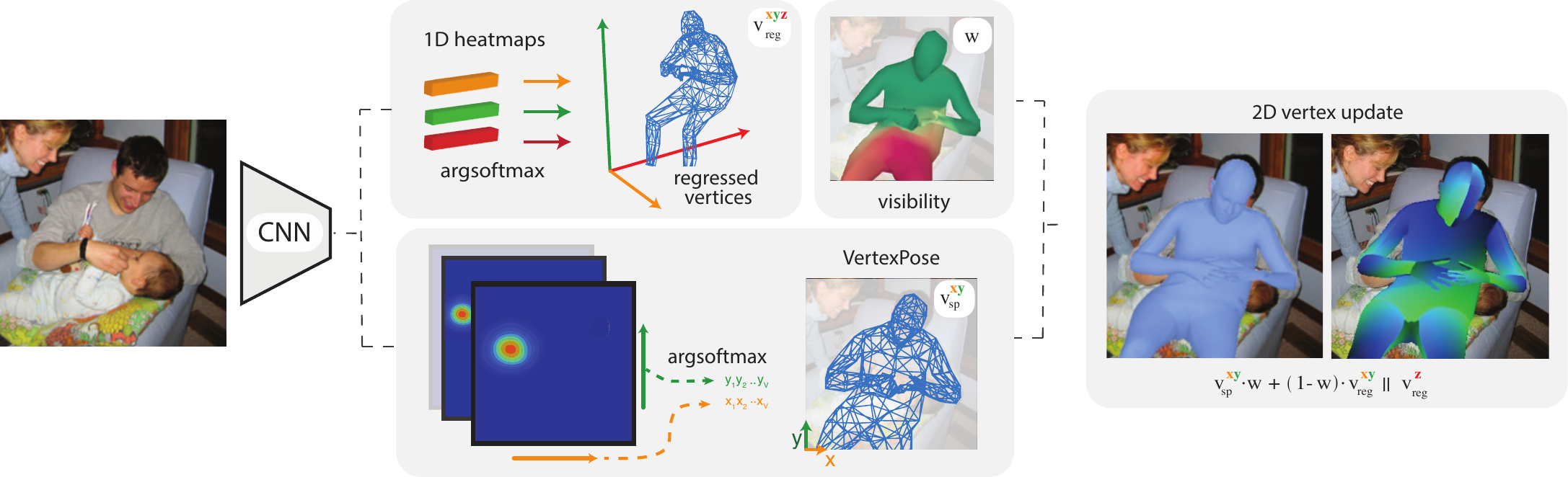}
    \caption{Meshpose Architecture: The lower \partpose branch extracts 
    multiple heatmaps from which, by applying the spatial argsoftmax operation, it computes precise $x$ and $y$ coordinates for all the vertices inside the input crop. The upper Regression branch computes the coordinates ($x$, $y$, and vertex depth $z$) for all vertices, along with their visibility scores $w$. The score $w$ will take lower values when the corresponding vertex is either occluded or fall outside the crop area. We differentiably combine the VertexPose and regressed coordinates via $w$ to get the final 3D mesh. We densely supervise the intermediate per-vertex heatmaps and the final output with UV, mesh and silhouette cues to end up with a low latency, image aligned, in-the-wild HMR system.
    }
    \label{fig:pipeline}
\end{figure*}

\section{Previous work}
Our starting point for this work is the understanding that Human Mesh Reconstruction systems are typically not grounded on pixel-level evidence for vertex positions, but instead try to predict them through the reconstruction of the much more complex structure of the body mesh. 
We take a bottom-up approach, where we first detect visible vertices through dense 2D heatmaps and then build the mesh around them. As we explain here, this has not been an obvious approach to HMR before our work. 

 Parametric 3D mesh reconstruction methods  such as \cite{kanazawa2018end,pavlakos2018learning,guler2019holopose,kolotouros2019learning,kocabas2020vibe,zanfir2020weakly,kocabas2021spec,kocabas2021pare,li2022cliff,kolotouros2021prohmr,choi2021beyond} methods provide rotation estimates to a forward kinematics (FK) recursion that unavoidably accumulates errors, thereby making it challenging to achieve good mesh alignment on wrists or ankles. Iterative fitting methods such as  \cite{guler2019holopose,kolotouros2019learning,song2020human,joo2020eft,pavlakos2019expressive,xu2020ghum} can mitigate this by minimizing the back-projection errors through gradient descent on the model parameters or by scaling up the required computational power \cite{goel2023humans}, but both are inappropriate for real-time inference. Variants of these works have been introduced to address additional complications due to 
human-human occlusion or object-human occlusion 
\cite{zhang2020object,Khirodkar_2022_CVPR,sun2021monocular,jiang2020mpshape,BEV,kocabas2021pare,muller2021self,hassan2019resolving} and perspective distortion effects \cite{li2022cliff,wang2023zolly} for in-the-wild scenes, where the problems described above become even harder. 

Inverse Kinematics (IK)-based methods \cite{iqbal2021kama,li2021hybrik,pliks,li2023niki}
are a step forward when it comes to localization accuracy in that they ensure that the recovered mesh `passes through' a set of 3D joints provided by a bottom-up system, yielding high 3D pose accuracy results. As our results show however, these methods fare poorly when it comes to 2D reprojection, since they are still constrained by the pose and shape variation of the employed parametric model. 

As an alternative to IK, recurrent refinement methods such as \cite{pymaf2021,refit,corona2022learned}
repeatedly estimate the positions of mesh vertices and sample features at the vertex positions to get a `second look' at the image. Our results indicate that their 2D reprojection accuracy is limited, while their recursive projection/lookup operations make them harder to deploy for mobile AR applications since they require custom layers which are not supported in CoreML\cite{coreml}/TFlite\cite{tflite} and end up being computational bottlenecks when performing inference on NPU/GPU-accelerated mobile devices.

Non-parametric HMR methods such as graph-convolutional \cite{kolotouros2019convolutional,choi2020pose2mesh}, heatmap-based \cite{moon2020i2l,yao2022learning,ma20233d}, or transformer-based models \cite{lin2021end,lin2021mesh,cho2022FastMETRO,PointHMR} bypass parametric models and directly regress the full body mesh. In principle this can avoid the problem of error accumulation during FK, yet as our DensePose results show in \reftable{tab:sota_3dpw} these methods still suffer when it comes to establishing accurate image reprojection. 
Our method bears similarity to recent methods for integral regression of per-vertex $x$/$y$/$z$ values \cite{moon2020i2l,yao2022learning} in that we
ground the vertex coordinates directly on image evidence, rather than regressing them through global mesh recovery systems. However we differ in that we directly connect our method to the DensePose estimation and training problems, thereby allowing us to get substantially better accuracy through direct optimization of the relevant objective. We use DensePose ground truth  as  weak supervision to localize our mesh vertices in 2D, and show that we can both deliver accurate 3D meshes and DensePose predictions. 

The DensePose task aims at associating every human pixel with its continuous, surface-based UV coordinates. This has been typically addressed through a dense regression task, where a CNN tries to directly recover these UV values with per-part UV regression heads, trained through a set of pixel-UV annotations. 
Unfortunately UV regression is of little direct value when it comes to mesh reconstruction: 2D vertex localization from DensePose requires multiple tricks (e.g. thresholding of UV distances, de-duplicating vertices, fixing left-right prediction errors for legs) and explains why this has not been adopted as a front-end processing for mesh recovery.

\section{Method}

In our work we recover a human body mesh from a single image 
in two stages as shown in \reffig{fig:pipeline}.
In the first stage, detailed in  \refsec{sparsepose}, we introduce ``\partposenogap'', a novel layer that serves a dual purpose: 
predicting DensePose and localizing mesh vertices in 2D. \partpose is designed to predict 2D heatmaps for a sparse set of body vertices that form a low-polygon approximation to a high-resolution mesh. Moreover, for each pixel, these heatmaps are integrated using barycentric combination, to compute the UV coordinates for any given pixel as DensePose.

We introduce novel  
losses to obtain weak supervision for \partpose heatmaps from DensePose data.
We show that even though we do not have direct supervision for the 2D locations of the \partpose vertices, we obtain  DensePose accuracy  comparable to  UV-based DensePose systems.

In the second stage, detailed in \refsec{meshpose}, we lift the 2D \partpose vertex positions to 3D, constructing a 3D mesh that accurately projects back to  \partpose vertices. 
We achieve this through a simple 1D integral regression task which estimates the root relative depth for all vertices in pixels.

We complement the \partposenogap-based losses with 3D counterparts that allow us to exploit 3D ground-truth
and pseudo ground-truth to jointly train \partpose and its associated 3D lifted predictions.

\begin{figure*}[!htb]
    \centering
    \begin{subfigure}[t]{0.4\textwidth}
    \includegraphics[width=\textwidth]{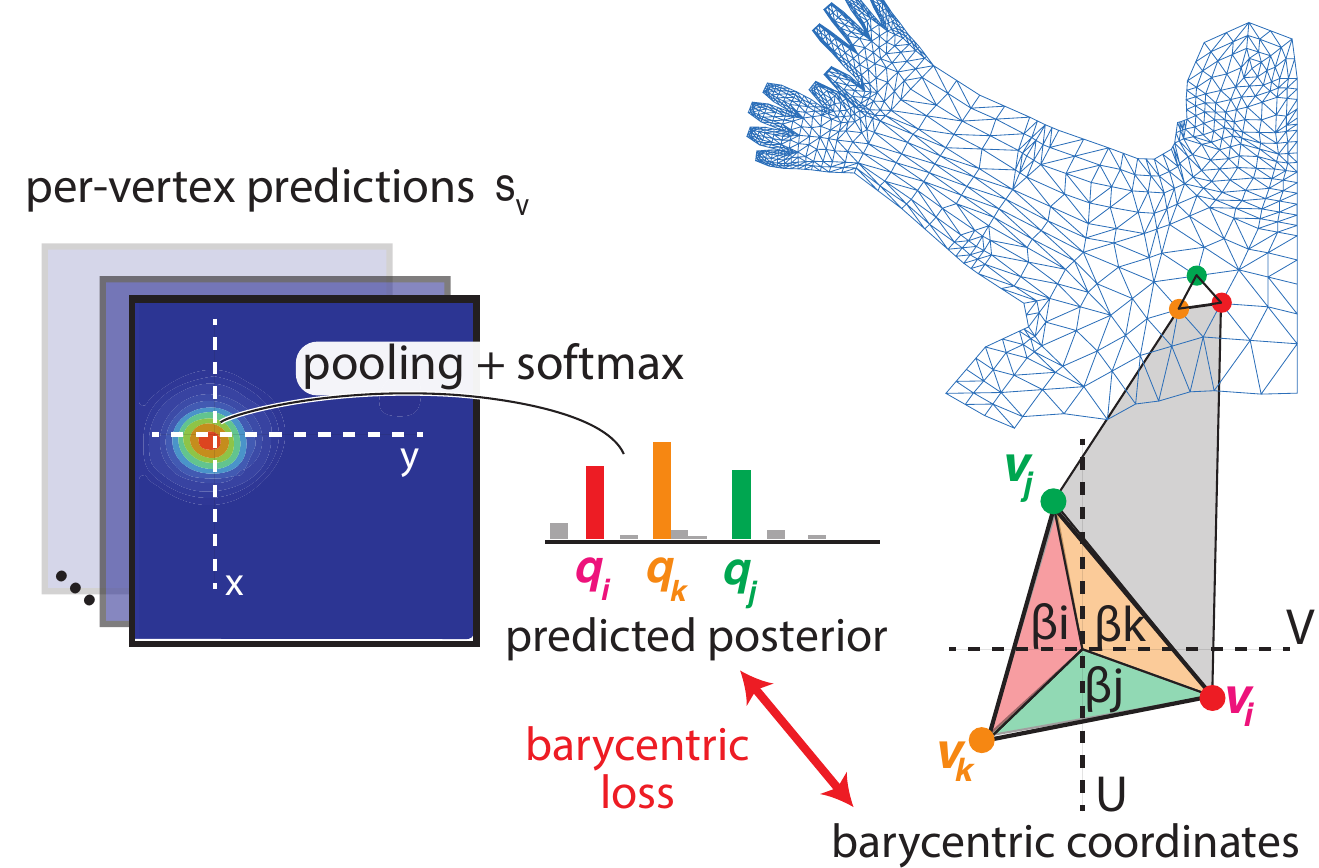}
    \caption{Barycentric loss}
    \end{subfigure}\hfill
    \begin{subfigure}[t]{0.55\textwidth}
    \includegraphics[width=\textwidth]{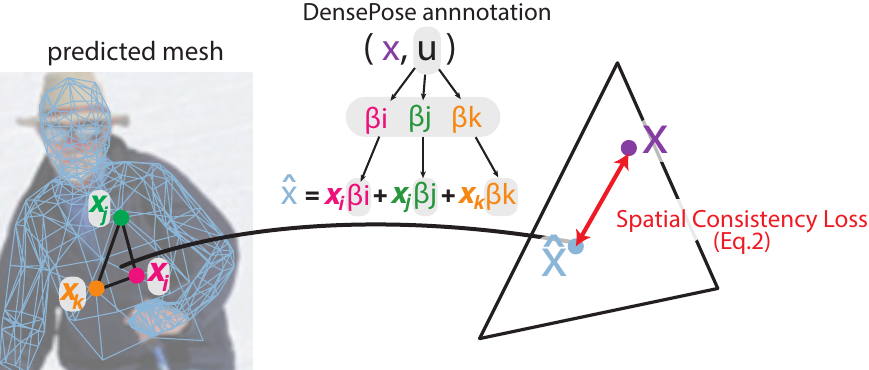}
    \caption{UV Consistency loss}
    \end{subfigure}
    \caption{
    Geometry-driven losses used to supervise \partpose with DensePose ground-truth.
    Our {\textit{barycentric loss}} requires that the per-pixel distribution over \partpose matches the UV annotation's barycentrics. 
    Our \textit{UV consistency loss} requires that the UV annotation's barycentrics at a labelled pixel $\pixel$ should  recover $\pixel$ based on a similar combination of \partpose vertices into $\hat{\pixel}$. 
    }
    \label{fig:2dlosses}
\end{figure*}
As the upper branch of our diagram shows, we augment the \partposenogap-based predictions with a per-vertex visibility estimate that is learned from weak supervision, as well as a global, image-level  3D regression of all  mesh vertices. This allows us to handle invisible parts by fusing the \partposenogap-based vertices with the latter prediction, which acts as a fallback. 

All stages rely on a single network that is trained end-to-end, but we separate their presentation and evaluation; we describe \partpose below and MeshPose in \refsec{meshpose}.

\subsection{\partposenogap: Vertex-based DensePose}

\label{sparsepose}

For \partpose we draw inspiration from the success of heatmap-based systems for 2D pose estimation e.g. \cite{xiao2018simple} and propose a similar layer to localize the low-poly vertices in 2D.
We start by describing the operation of the \partpose layer and then introduce new losses that allow  training it from DensePose ground-truth through weak supervision. 

\subsubsection{\partpose layer}
The \partpose layer consists of an $H\times W \times V$  tensor $\sparsepose$ where $H, W$ are tensor height/width and $V$ is the number of vertices.
This provides for each mesh  vertex the score of all 2D input positions, yielding a set of heatmaps that serve
 both 2D vertex localization and dense UV prediction. 
We further denote by $\sparseposepix = \sparsepose[x,y,:]$ the $1\times V$ set of \partpose scores at a given position $\pixel = (x, y)$ 
and by $\sparseposechan_v = \sparsepose[:,:,v]$ the $H\times W$ heatmap corresponding to a given vertex $v$.

We localize every vertex $v$ as the argsoftmax of $\sparseposechan_v$:
\begin{equation}
\pixel_v = \mathrm{argsoftmax}(\sparseposechan_v) = \frac{\sum_{i} \pixel_i \exp(\alpha \sparseposechan_v[\pixel_i])}{\sum_{i} \exp(\alpha \sparseposechan_v[\pixel_i])},
\label{eq:argsoftmax}
\end{equation}
where the exponentiation and normalization turns the possibly negative heatmaps into a distribution over positions and
the scaling parameter $\alpha$ allows us to make the resulting distribution more peaked and helps with localization.
 
\newcommand{\face}{f}
\newcommand{\faces}{\mathcal{F}}

We estimate the $UV$ value at a position $\pixel$ based on the per-pixel posterior over vertices  $q_v = \frac{\exp(\sparseposepix_v)}{\sum_{k=1}^{V}{\exp(\sparseposepix_k)}}$. We first identify the 
mesh face $\face={i,j,k}$ whose vertices have
the largest cumulative score: $\face^{\ast} = \mathrm{argmax}_{\face\in \faces} \sum_{n \in \face}q_n$.
We then estimate the $UV$ value as the barycentric combination of the UV values of the face vertices $\uv_m, m \in \face^{\ast}$:
\begin{equation}
\uv = \sum_{m \in \face^{\ast}} \beta_m \uv_m, ~\mathrm{where}~ \beta_m = \frac{q_m}{\sum_{n \in \face^{\ast}} q_n}
\label{eq:baryinterp}
\end{equation}
where we treat the normalized posterior over vertices that form $\face^{\ast}$ as an estimate of the pixel's barycentric coordinates.   

We note  the dual nature of our \partpose layer: UV estimation is not needed at inference time for our network, but is rather used as a means to supervising our network through DensePose ground-truth.     By contrast vertex localization cannot be directly supervised, but is the  2D substrate for our 3D MeshPose system. 

VertexPose additionally generates a segmentation mask to accurately localize the foreground, which is a required component for the evaluation of DensePose metrics for this subsystem. This layer is omitted from the final HMR system.

\subsubsection{\partpose training}

The main challenge when training the \partpose layer is the absence of direct supervision for the 2D \partpose vertex positions: the DensePose dataset was collected with continuous regression in mind and relied on annotating random body pixels with their associated UV values. This means that we do not have strong supervision at the level of per-vertex 2D ground-truth locations, which would allow us to directly also use the loss functions used for 2D pose estimation e.g. in \cite{xiao2018simple}. We mitigate this by introducing novel losses that exploit the underlying geometric nature of \partpose and thereby allow us to use DensePose data for weak supervision.

\newcommand{\mypart}[1]{\noindent\textbf{#1:}}
\mypart{Barycentric Cross-Entropy Loss}

This loss forces the softmax-based posterior over \partpose vertices to approximate the barycentric coordinates on any pixel that has UV annotation, as shown in \reffig{fig:2dlosses}(a).

In particular for any pixel $\pixel = (x, y)$ that comes with a DensePose annotation  with UV coordinates $\uv$ we introduce a loss on the  \partpose scores $\sparseposepix = \sparsepose[x,y,:]$ at that pixel. 
We phrase the task as one of competition among the \partpose vertices for the occupancy of the particular pixel.
If a vertex $v$ landed precisely on a given pixel we could impose at that point a standard Cross-Entropy loss using the one-hot encoding of that vertex. But this is unlikely to happen, since DensePose ground-truth was originally not sampled on specific landmark locations such as the vertices.

Instead we form our loss by interpreting the ground-truth barycentric coordinates  as a discrete distribution on   vertices of the ground-truth triangle $\face$. We use this to penalize the softmax-based posterior using the general definition of the cross-entropy loss: 
\vspace{-2mm}
\begin{eqnarray}
\mathcal{L}_{BL} = -\sum_v p_v \log(q_v),
\label{eq:cross_entropy} 
~ \mathrm{with}~
p_v = \left\{\begin{array}{cc} \beta_v\! &\! v \in \face \\ 0 \!&\! v \notin \face
\end{array}\right.
\end{eqnarray}
where we replace the common one-hot encoding of the correct label with a distribution on vertices, $p_v$, forcing the \partposenogap-based posterior $\mathbf{q}$ to align with the barycentric-based distribution $\mathbf{p}$.
Our results indicate the advantage of using this geometry-inspired loss instead of a cruder, nearest neighbor assignment of annotated pixels to their nearest mesh vertex. 
 
\mypart{UV Consistency Loss}
\label{uvloss}
This loss forces \partpose to place vertices so that the image coordinates of annotated UV values align with the positions where they were annotated, as shown in \reffig{fig:2dlosses}(b).

In particular we turn a pixel's DensePose annotation $(\pixel,\uv)$  into a constraint on the \partpose heatmaps $\sparseposechan_v = \sparsepose[:,:,v], v \in (i,j,k)$ of the three vertices $(i,j,k)$ used to compute the barycentric coordinates $(\beta_i,\beta_j,\beta_k)$ of $\uv$. 
These three barycentric coordinates allow us to localize the pixel's corresponding point on the 3D surface as a convex combination of these three vertices. When projected to the image this relationship should still roughly hold, modulo depth-based perspective distortion effects, which we consider  negligible within a triangle. 
Our loss enforces this: when combining  the estimated 2D positions of the three vertices $\pixel_v = \mathrm{argsoftmax}(\sparseposechan_v), v \in \{i,j,k\}$, with barycentric weights $\beta_v$, we should be able to recover the position of $\pixel$:
\vspace{-3mm}
\begin{gather}
L_{\mathrm{consistency}} = \|\pixel - \hat{\pixel}\|,~~\mathrm{where}~~
\hat{\pixel} = \sum_{v \in \{i,j,k\}} \beta_v \pixel_v
\end{gather}

This forces the heatmap $\sparseposechan_v$ to properly localize $\pixel_v$.  without direct supervision for vertex $v$.

Both of these losses are efficient to evaluate and as our results in \refsec{results} show, they add up to the training of a \partpose system 
that even outperforms the UV-based DensePose baseline when trained with identical data and experimental settings. Still, we consider the competition with UV-based DensePose systems to be of secondary importance compared to being able to directly predict a mesh based on the subsequent lifting of the  estimated vertices to 3D, as described in \refsec{meshpose}. 
\mycomment{
\subsection{Part Segmentation Loss}
\label{partloss}

\input{Figures/part_segmentation}

Our last source of \partpose supervision relies on the dense part segmentation maps provided in the DensePose ground-truth.  For every visible body pixel $\pixel$ we know its part label $l$ and for every body part we know that set over vertices that it comprises $\mathcal{S}_l$. We use this to obtain a part-level posterior distribution:
\ba 
p_l = \sum_{v \in \mathcal{S}_l} q_v,\quad l \in 1,\ldots,P
\ea
by collapsing the \partposenogap-level softmax posterior $q_v$ (defined in 
\refeq{eq:cross_entropy}). This posterior distribution is supervised based on the standard Cross-Entropy loss obtained from the one-hot encoding of the part label. 
}

\subsection{MeshPose: Lifting \partpose to 3D}
\newcommand{\id}{d}
\newcommand{\Id}{D}
\label{meshpose}

Having outlined our method to localize mesh vertices in 2D we now turn to converting them into a 3D mesh. As shown in \reffig{fig:pipeline}, our method consists of retaining the image localization information of \partpose where available and filling in the remaining information by values regressed by a separate network branch. 

Inspired by \cite{pliks,moon2020i2l} we take the backbone CNN's last tensor, average pool it and transform the result through 1D convolutional layers to regress a $4 \times  64 \times V$ tensor, where $V$ is the number of the low-poly vertices, the four channels correspond to $X,Y,Z$ values and a per-vertex visibility label $w$ and $64$ are the number of bins used for argsoftmax voting. In particular for every regressed vertex $V_{reg}^{XYZ}$ its $X,Y,Z$ values are obtained separately per dimension by applying 1D argsoftmax  while the visibility $w$ is obtained by mean pooling followed by a sigmoid unit. We detail how we supervise those terms below.

\vspace{-0.5mm}
\subsubsection{Visibility prediction}
The visibility label dictates on a per-vertex level whether we should rely on the \partposenogap-based 2D position, $V_{sp}^{XY}$ or fall back to the $V_{reg}^{XY}$ value regressed at this stage. This allows us to accommodate occluded areas, or tight crops that omit part of the human body, as is regularly the case for selfie images. The  2D location of a MeshPose vertex is their visibility-weighted average: $V_{mp}^{XY} =  V_{sp}^{XY}  w + V_{reg}^{XY} (1-w)$. This differentiable expression allows us to estimate visibility  through end-to-end back-propagation,  but we also use two additional methods for visibility supervision.

Firstly we estimate partial vertex visibility based on the available ground-truth: for any $(\pixel,\uv)$ annotation pair contained in the DensePose dataset, we declare as visible all three vertices that lie on the mesh triangle containing $\uv$. We also declare as non-visible every vertex where the mesh supervision (obtained from \cite{NeuralAnnot}) is outside the image crop. For such vertices we can supervise visibility based on a standard binary cross-entropy loss. 

Secondly, we also supervise visibility at the mesh level. For this we use differentiable rendering with the per-vertex texture set to equal the predicted visibility label. This produces a soft visibility mask,  shown also in \reffig{fig:pipeline} as a heatmap, which indicates the image area that is covered by the person's body. This can be supervised at the region level based on the DensePose dataset's instance segmentation masks, using a mix of an $\ell_2$ loss with the integral boundary loss introduced in \cite{ibl}.        
These two sources of visibility supervision gave substantial improvements as also shown by our results.

\subsubsection{Depth regression}
We adopt a weak perspective camera model as in \cite{moon2020i2l,yao2022learning} and consider that each vertex lies on a ray that crosses the image plane at the \partposenogap-based 2D position. We thereby limit 3D lifting to the task of estimating the vertex depth on that ray.
Rather than directly regress the depth of a vertex, we predict its  depth  relative to the `root' of the mesh (sternum). The latter is predicted by a separate RootNet network \cite{moon2019camera}, leaving to our network the task of relative depth estimation. 
We estimate depth in pixel units, based on the same rigid transform used to associate the metric joint (x,y) positions with their 2D pixel counterparts.

\subsubsection{MeshPose output}
The  MeshPose 3D prediction concatenates the visibility-weighted 2D location with the depth prediction:
\begin{equation}
V_{mp}^{XYZ} =  (V_{sp}^{XY}  w + V_{reg}^{XY} (1-w) || V_{reg}^{Z})
\end{equation}
All terms  are differentiable, allowing us to train our network end-to-end based on 3D mesh supervision. 

We note that we can optionally also transform the resulting low-poly mesh  into a high-poly counterpart through an MLP-based upsampling as in \cite{lin2021end}. We have trained such an MLP and used it both for mesh visualizations and performance evaluations. Even though it primarily serves as a ``visual embelishment'' of the low-poly prediction, it also provides some form of regularization when trained with noised low-poly inputs. 

Finally, if a parametric representation is needed for an application the MeshPose prediction lends itself easily to Inverse Kinematics-based processing \cite{li2021hybrik} by using landmarker-based 3D joint estimates. Our implementation of HybridIK-type decoding yields virtually identical 3D metrics, but comes at the cost of a drop in DensePose accuracy, as expected, hence we omit it from evaluations.

\subsubsection{3D supervision}
\newcommand{\Pose}{\mathbf{X}}

In order to supervise the 3D coordinates of the low poly mesh we use motion capture ground truth (GT) meshes, weak supervision for the in-the-wild COCO dataset (pseudo-GT), and the losses described below.

\mypart{Vertex Localization, Edge and Normal Loss} \label{edgeloss} The position of our 3D vertices can be directly compared to the (pseudo) GT in terms of an L2 loss (`localization loss'), while we can also penalize the distortion of the edge lengths between two adjacent vertices (`edge loss').
We note that the second loss does not necessarily guarantee good alignment to the image, but ensures we do not arbitrarily stretch or shrink the mesh to reduce other losses. To further reduce mesh curvature artifacts we use a third, ('normal cosine loss') that penalizes the deviation of predicted vertex normals.

\mypart{Joint Localization Loss} The last form of supervision relies on standard 2D or 3D joint GT that is more readily available through image annotations or motion capture, respectively. 

The position of each joint is estimated as a weighted average of a subset of nearby mesh vertices, implemented as a precomputed linear regression (landmarker).

We compare the predicted joints to the GT using its full 3D coordinates or their projections on the image, based on whether we have 3D or 2D supervision. 
Even though only a sparse subset of vertices contribute to the prediction of any joint,  the edge loss described above helps diffuse the supervision to the remainder of the mesh.

\section{Results}
\label{results}

We start by describing experimental settings, and then proceed with quantitative and qualitative evaluation. \textit{Due to lack of space we provide additional ablations and experimental results in the Supplemental Material, including results on videos. }

\subsection{Datasets}

We use the manual DensePose annotations provided in \cite{guler2018densepose} on the \textbf{MS-COCO} \cite{lin2014microsoft} dataset for training the \partpose system. 
For 3D joint supervision we use three  datasets: 
(i) The \textbf{Human3.6M} \cite{ionescu2013human3}: MoCap dataset and follow the protocol of  \cite{kolotouros2019learning} (subjects (S1, S5, S6, S7, S8) for training)  
(ii) The \noindent \textbf{MPI-INF-3DHP} \cite{mehta2017monocular}: Multi-view  dataset, following the train split of \cite{kolotouros2019learning}.
(iii)  The \textbf{3DPW} \cite{von2018recovering} in-the-wild outdoor benchmark for 3D pose and shape estimation containing 3D annotations from IMU devices.
Furthermore, we augment the MS-COCO dataset with the 3D mesh pseudo GT annotations of \cite{NeuralAnnot} for vertex-level mesh supervision.

\subsection{Evaluation Metrics}
We adopt the evaluation framework outlined in \cite{guler2018densepose} for the DensePose task, where we report on two key metrics: \textbf{AP} (Average Precision) and \textbf{AR} (Average Recall). These metrics quantify the accuracy of the dense correspondences from UV coordinate predictions on images. For mesh recovery methods this can be interpreted as measuring mesh alignment accuracy after projection. We measure the correspondence accuracy by rendering UV coordinates of the visible mesh surface. In addition to these 2D metrics, we evaluate the accuracy of 2D COCO keypoints prediction, quantified through Average Precision and Recall. This evaluation is conducted both across all instances (\textbf{AP-All} and \textbf{AR-All}), as well as only on instances where at least 80\% of the keypoints are visible (\textbf{AP-80\%} and \textbf{AR-80\%}). All 2D metrics are evaluated on DensePose-COCO, a subset of COCO~\cite{lin2014microsoft} introduced in~\cite{guler2018densepose}.

For 3D pose evaluation we employ a landmarker on top of the high poly 3D mesh to compute the 14 LSP joints for the evaluation on 3DPW dataset  
\cite{kanazawa2018end, kolotouros2019learning}. Then we compute the Euclidean distances (in millimeter (mm)) of 3D points between the predictions and GT as described by the following metrics:
(i)  \textbf{MPJPE} (Mean Per Joint Position Error) first aligns the predicted and GT 3D joints at the 3D position of the pelvis, evaluating the predicted pose by taking into account the global rotation. 
(ii) \textbf{PA-MPJPE} (Procrustes-Aligned Mean Per Joint Position Error, or reconstruction error) performs Procrustes alignment before computing MPJPE, eliminating any error due to wrong global scale and global rotation.
(iii) \textbf{PVE} (Per Vertex Error) does the same alignment as MPJPE and then calculates the distances of vertices of human mesh, evaluating also the mesh shape additionally to the 3D skeleton pose. 

\newcommand{\fnl}[1]{#1}
\subsection{\partpose evaluation}

We start by examining the impact of training with the vertex-based (\partposenogap) approach compared to 
the UV-regression based (DensePose) representations for the DensePose task. To keep the comparison apples-to-apples in
\reftable{table:dp_vs_sp} we compare experiments with identical backbones and training settings, where we closely follow \cite{guler2018densepose} for designing the DensePose baseline.  We observe that the \partposenogap-based results compare favorably to their DensePose-based counterparts, confirming the validity of the proposed approach. 

In \reftable{table:decoding} we analyze the impact of the barycentric interpolation strategy used to predict UV in \refeq{eq:baryinterp}. We compare it to simpler baselines of using the UV of the strongest vertex at any pixel (`Nearest') or doing argsoftmax over all vertices rather than those of the strongest triangle (`Global Average'). The results indicate the merit of the smooth transition between vertices secured by barycentric interpolation.

We consider improvements in the 2D DensePose task to be of secondary importance compared to improving the 3D HMR's DensePose performance, hence keep the remaining results focused on MeshPose.

\newcolumntype{Z}{>{\setbox0=\hbox\bgroup}c<{\egroup}@{}}
\begin{table}[!htp]\centering
\begin{subtable}[c]{\columnwidth}
\centering
\scalebox{0.8}{
\begin{tabular}{cH|ccHHH|ccHHH}
  \toprule
   & \hspace*{\fill} architecture \hspace*{\fill} & $\textbf{AP}$ & $\textbf{AP}_{50}$ & $\textbf{AP}_{75}$ & $\textbf{AP}_{M}$ & $\textbf{AP}_{L}$ & $\textbf{AR}$ &
   $\textbf{AR}_{50}$ & $\textbf{AR}_{75}$ & $\textbf{AR}_{M}$ &
   $\textbf{AR}_{L}$\\
   \midrule

DP - MBNet &No mask &53.54 &91.66 &58.29 &54.75 &53.65 &64.09 &95.68 &75.39 &54.89 &64.73 \\
SP - MBNet &No mask &
54.08	& 93.01 &	59.41	& 55.94	& 54.32	& 64.46& 	96.03	& 76.10	& 56.24	& 65.03
\\\midrule
DP - ResNet &No mask &57.31 &93.06 &65.60 &57.72 &57.49 &67.51 &96.30 &80.29 &57.87 &68.17 \\
SP - ResNet &No mask & 
58.87&	93.05&	68.34&	60.10&	58.90&	68.73&	96.26&	82.08&	60.14&	69.33 \\
\midrule
DP - HRNet32 &No mask &61.12 &94.84 &72.23 &61.39 &61.41 &70.53 &97.33 &84.26 &61.56 &71.16 \\
SP - HRNet32 &No mask & 
61.24	&94.63&	72.61&	61.68&	61.54&	70.52&	97.10&	84.53&	61.77&	71.13
 \\\midrule
DP - HRNet48 &No mask &62.74 &95.04 &75.39 &63.86 &63.03 &71.95 &97.50 &86.18 &64.11 &72.49 \\
SP - HRNet48 &No mask &
63.32	&95.12&	76.48	&64.75	&63.63& 72.14&	97.73&	87.07&	64.89&	72.65\\
\bottomrule
\end{tabular}
}
\caption{\partpose vs DensePose.}\label{table:dp_vs_sp}
\end{subtable}
\hfill
\begin{subtable}[c]{\columnwidth}
\centering

\scalebox{0.8}{
 \begin{tabular}{cHccHHHccHHH}
  \toprule
   & \hspace*{\fill} architecture \hspace*{\fill} & $\textbf{AP}$ & $\textbf{AP}_{50}$ & $\textbf{AP}_{75}$ & $\textbf{AP}_{M}$ & $\textbf{AP}_{L}$ & $\textbf{AR}$ &
   $\textbf{AR}_{50}$ & $\textbf{AR}_{75}$ & $\textbf{AR}_{M}$ &
   $\textbf{AR}_{L}$\\
   \midrule

 Barycentric &No mask & 
61.24	&94.63&	72.61&	61.68&	61.54&	70.52&	97.10&	84.53&	61.77&	71.13
 \\
Closest &No mask &59.69 &95.10 &70.08 &61.39 &60.01 &68.80 &97.59 &83.24 &61.49 &69.31 \\
Global Average &No mask &33.35 &89.58 &11.00 &37.13 &33.61 &43.23 &94.38 &31.79 &37.16 &43.64 \\

\bottomrule
\end{tabular}
}
\caption{UV aggregation strategy}\label{table:decoding}
\end{subtable}
\caption{
Analysis of \partpose performance on the DensePose-COCO dataset. We evaluate the impact of the backbone choice and the UV decoding strategy.
}
\vspace{-3mm}
\end{table}

\begin{figure}
    \includegraphics[width=\linewidth]{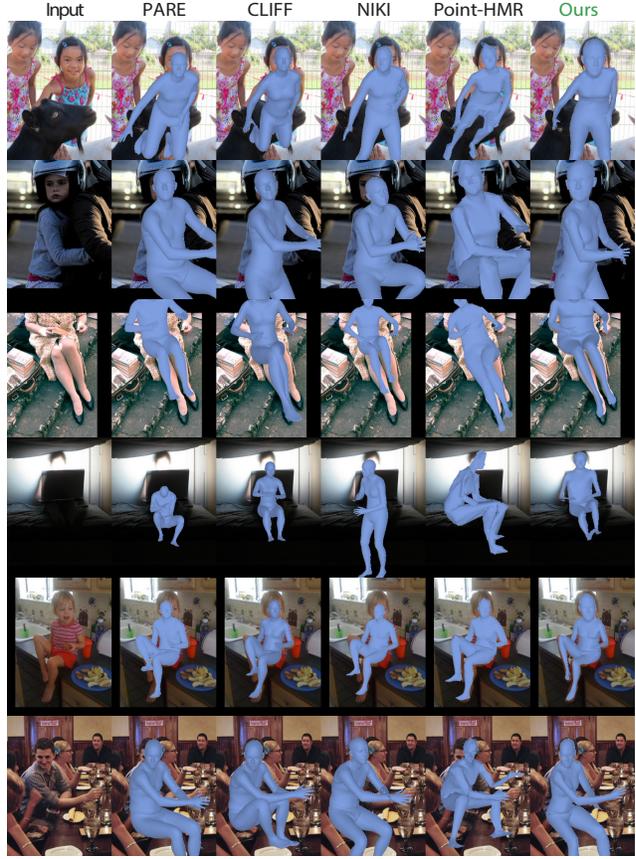} 
    \caption{Qualitative comparison on COCO against 4 state-of-the-art mesh reconstruction systems. MeshPose is robust to severe occlusions, partial body cropping and body shapes.}
    \label{fig:COCO_comparison}
\end{figure}

\subsection{MeshPose evaluation}

In  \reftable{tab:ablation_sp2mp} we start by ablating the impact of the types of supervision used for our full MeshPose system.
 Starting from only 3D losses (vertex localization, edge and normal loss - $\mathcal{L}_{3D}$) on row 1, we show the impact of  adding 2D losses (barycentric and uv consistency loss - $\mathcal{L}_{2D}$) and visibility supervision (partial visibility and rendering loss - $\mathcal{L}_{W}$). We show that adding 2D losses leads to a moderate drop in 3D accuracy but boosts DensePose reprojection metrics, while adding visibility supervision helps improve both tasks at the same time - effectively helping them coexist.

\begin{table}[htbp]
  \centering
  \scalebox{0.68}{
    \begin{tabular}{|HcccHHH|ccc|ccc|}
    \toprule
    \multirow{2}[2]{*}{Id} & \multicolumn{6}{c|}{Losses}  & \multicolumn{3}{c|}{3DPW} & \multicolumn{3}{c|}{COCO-DensePose} \\
          & $\mathcal{L}_{3D}$ & $\mathcal{L}_{2D} $ & $\mathcal{L}_{W}$ & $\mathcal{L}_{\text{rend}}$ & merging & 3DPW  & MPJPE & PA-MPJPE & PVE & AP    & AR  & IoU \\
    \midrule
    1 & \checkmark & & & & \checkmark & \checkmark & \textbf{76.52} & \textbf{45.31} & \textbf{91.88} & 38.22 & 49.07 & 56.56 \\
    2 & \checkmark & \checkmark &  & \checkmark & & \checkmark & 81.37 & 50.40 & 101.68 & 44.09 & 53.31 & 57.64 \\    

    3 & \checkmark & \checkmark & \checkmark & \checkmark & \checkmark & \checkmark & 77.10 & 46.54 & 94.78 & \textbf{45.51} & \textbf{55.18} & \textbf{60.24}\\

    \bottomrule
    \end{tabular}
  }
  \vspace{-0.1cm}
  \caption{Ablation table evaluated in terms of 3D metrics (3DPW) and 2D reprojection accuracy (COCO-DensePose).}
  \label{tab:ablation_sp2mp}
\end{table}

Turning to comparisons with HMR methods, in (\reftable{tab:sota_3dpw}) we extensively compare our approach with 10 other SOTA architectures in terms of efficiency (measured in \#Parameters and FPS) and accuracy. We report the performance on 3 tasks, (i) 2D DensePose-COCO Keypoints, (ii) DensePose and (iii) 3D alignment on 3DPW and Human 3.6M.

MeshPose is by far the most efficient approach for achieving real-time prediction while getting the best 2D reprojection performance. This large improvement comes at the cost of a small impact on 3D metrics. 

\newcommand{\threedpw}{3DPW~}

\begin{table*}
\centering
\resizebox{\textwidth}{!}{
    \begin{tabular}{ll|r|r|r|r|r|r|r|r|r|r|r|r|r}
        \toprule
        & \bf Method & \multicolumn{2}{c}{\bf Efficiency} & \multicolumn{4}{|c}{\bf 2D COCO KeyPoints $\uparrow$} & \multicolumn{2}{|c}{\bf DensePose $\uparrow$} & \multicolumn{3}{|c}{\bf 3DPW $\downarrow$} & \multicolumn{2}{|c}{\bf Human 3.6M $\downarrow$} \\
        \cmidrule(lr){3-15}
        &  & \#Params $\downarrow$ & FPS $\uparrow$ & AP-All  & AR-All  & AP-80\%  & AR-80\%  & AP  & AR  & MPJPE  & PAMPJPE  & PVE  & MPJPE  & PAMPJPE  \\ 
        \midrule
        \parbox[t]{2mm}{\multirow{6}{*}{\rotatebox[origin=c]{90}{param}}} 
            & DaNet~\cite{zhang2020densepose2smpl} & 102.25 & 11.95 & 19.80 & 36.20 & 52.60 & 65.20 & 16.42 & 29.24 & 85.50 & 54.80 & 110.80 & 54.60 & 42.90\\
            & HybrIK~\cite{li2021hybrik} & 75.69 & 7.57 & 10.60 & 24.90 & 31.70 & 47.50 & 20.85 & 34.44 & 71.60 & 41.80 & 82.30 & \textbf{47.00} & \textbf{29.80}\\
            & PARE~\cite{kocabas2021pare} & 32.86 & 20.48 & 33.20 & 48.10 & 56.90 & 67.20 & 30.02 & 41.54 & 74.50 & 46.50 & 88.60 & - & -\\
            & CLIFF~\cite{li2022cliff} & 78.89 & 22.42 & 33.20 & 49.30 & 61.40 & 72.50 & 30.99 & 41.83 & \textbf{69.00} & 43.00 & \textbf{81.20} & 47.10 & 32.70\\

            & PyMaf~\cite{pymaf2021} & 45.18 & 35.57 & 23.00 & 40.40 & 58.00 & 70.00 & 17.62 & 30.69 & 92.80 & 58.90 & 110.10 & 57.70 & 40.50\\
            & NIKI~\cite{li2023niki} & 92.17 & 6.18 & 21.30 & 37.20 & 48.50 & 61.50 & 25.11 & 37.39 & 71.70 & \textbf{41.00} & 86.90 & - & -\\
        \midrule
        \parbox[t]{2mm}{\multirow{4}{*}{\rotatebox[origin=c]{90}{n-param}}}
            & Metro~\cite{lin2021end} & 243.07 & 15.47 & 9.50 & 22.40 & 30.40 & 45.10 & 9.28 & 21.16 & 77.10 & 47.90 & 88.20 & 54.40 & 34.50\\
            & Graphormer~\cite{lin2021mesh} & 226.21 & 14.42 & 12.00 & 25.70 & 35.90 & 49.50 & 13.57 & 26.20 & 74.70 & 45.60 & 87.70 & 51.20 & 34.50\\
            & FastMetro~\cite{cho2022FastMETRO} & 153.74 & 16.04 & 13.60 & 28.00 & 39.30 & 53.20 & 13.76 & 26.21 & 73.50 & 44.60 & 84.10 & 52.20 & 33.70\\
            & PointHMR~\cite{PointHMR} & 59.09 & 15.38 & 17.70 & 32.70 & 44.40 & 57.30 & 19.58 & 32.18 & 73.90 & 44.90 & 85.50 & 48.30 & 32.90\\
        \midrule
        \parbox[t]{2mm}{\multirow{3}{*}{\rotatebox[origin=c]{90}{ours}}}
            & MeshPose (HRNet32 \cite{wang2020deep}) & 46.29 & 28.66 & \textbf{47.30} & \textbf{61.30} & \textbf{71.20} & \textbf{79.60} & \textbf{47.87} & \textbf{57.62} & 76.08 & 46.73 & 92.70 & 50.76 & 35.37\\    
            & MeshPoseS (ResNet50 \cite{he2016deep}) & 45.37 & \textbf{124.28} & 43.80 & 58.10 & 67.00 & 76.50 & 44.41 & 54.49 & 80.03 & 48.97 & 97.96 & 56.33 & 37.64\\
            & MeshPoseXS (MBNet140 \cite{mbnet}) & \textbf{21.25} & 124.21 & 40.60 & 55.20 & 63.60 & 73.90 & 38.56 & 49.20 & 79.15 & 49.71 & 96.49 & 58.40 & 41.63\\
        \bottomrule
    \end{tabular}
}
\caption{Evaluation of network efficiency, 2D accuracy in COCO-DensePose and 3D errors in the 3DPW and Human3.6M datasets. The variants of our system achieve superior performance in 2D metrics (2D Keypoints, Densepose) when compared to other methods, while they achieve comparable 3D accuracy. At the same time they are substantially more efficient in terms of FPS and \# of parameters.} 
 
    \label{tab:sota_3dpw}
    \vspace{-2mm}
\end{table*}

Our method achieves competitive performance against other parametric and non-parametric approaches, while largely outperforming them on the DensePose task. More specifically, MeshPose achieves much better DensePose metrics compared against CLIFF and NIKI methods, which have the best scores for the HMR task.

\begin{figure}
    \includegraphics[width=0.5\textwidth]{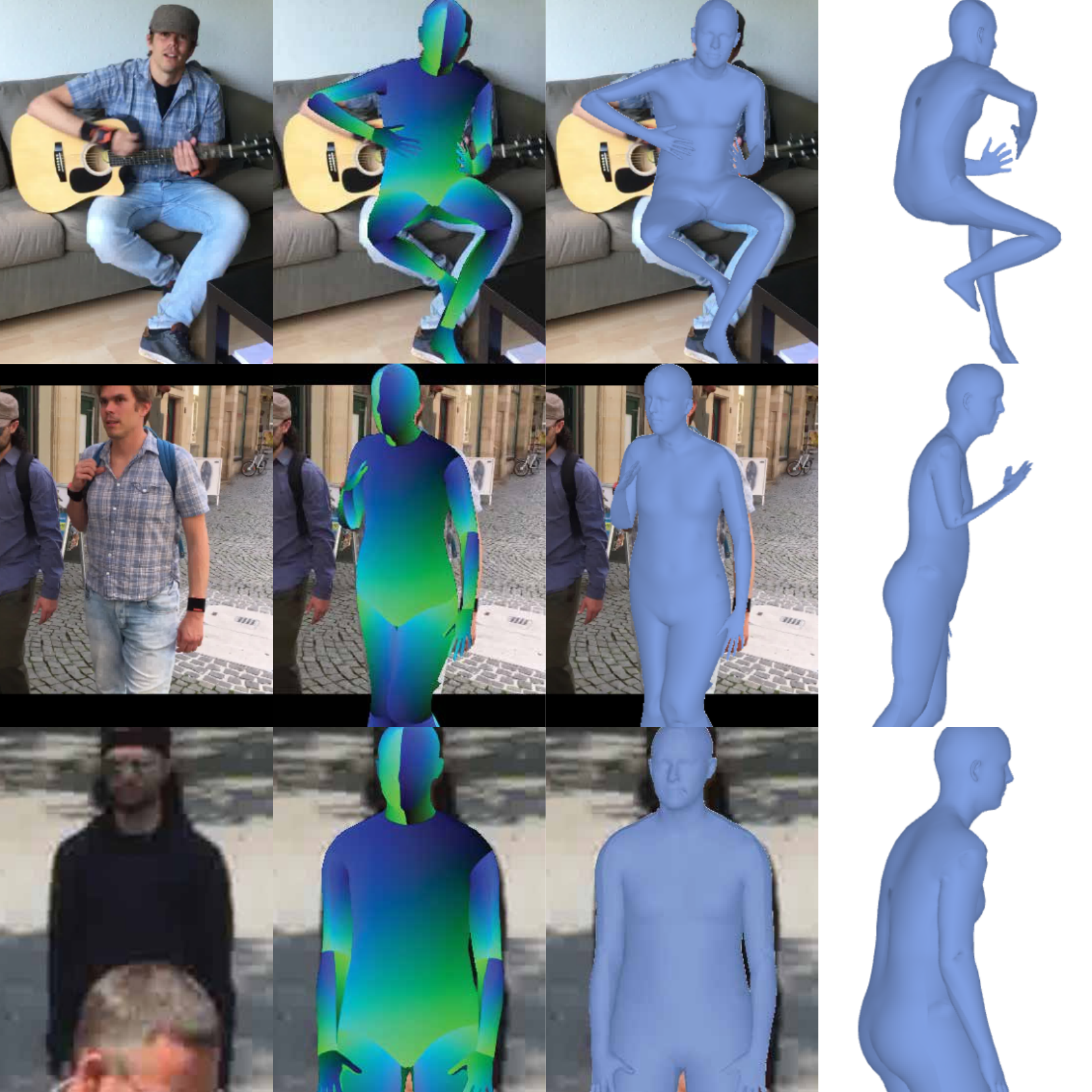} 
    \caption{Qualitative results on 3DPW on front and side views. Our method shows strong 2D alignment with accurate 3D mesh prediction.}
    \label{fig:3DPW_uv_front_side}
    
\end{figure}

Fig.~\ref{fig:COCO_comparison} depicts some examples comparing the proposed MeshPose against the PARE, CLIFF, NIKI and Point-HMR methods. We can see that MeshPose produces meshes that, when projected on the image, align much better than competing methods. Other methods fail e.g. when reconstructing children, when a large part of the body is occluded and have inferior alignment around the limbs. In Figure~\ref{fig:3DPW_uv_front_side}, we further demonstrate the performance of our system in terms of mesh reconstruction by including side views of our meshes predicted on 3DPW images.

\begin{figure}[h]
    \includegraphics[width=\linewidth]{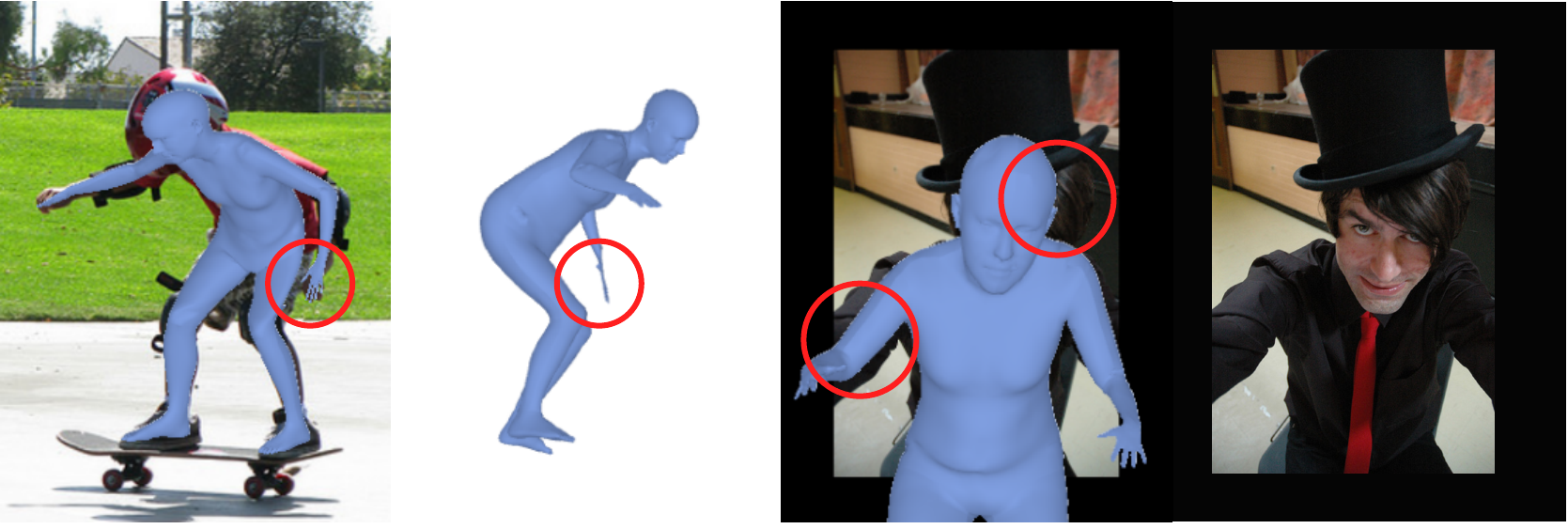} 
    \caption{Typical failure cases of MeshPose include a hand flattening artifact and imperfect image alignment in the presence of perspective distortion.}
    \label{fig:failure_cases}
\end{figure}

\subsection{Limitations}
\label{sec:limitations}
As shown on Figure~\ref{fig:failure_cases}, the two main failure modes of our system are hand/arm flattening artifact and imperfect alignment for perspectively distorted inputs. The flattening artifact primarily arises due to the mesh upsampler's lack of training on examples with articulated hands. The imperfect perspective alignment is caused by our system's reliance on the assumption of weak perspective camera model.
Still, the robustness of our method is comparable to DensePose, hence we do not have catastrophic failures (e.g wrong torso pose) in the presence of heavy occlusions.

\section{Conclusion}
\label{sec:conclusion}
In this work, we started by observing the limited 2D reprojection accuracy of current HMR systems, which limits their applicability to augmented reality applications - e.g. virtual try-on for garments and accessories. To address this we have introduced MeshPose, a system that
bridges the DensePose and HMR problems and  substantially improves the image reprojection accuracy of HMR, while retaining accurate 3D pose. Beyond improved accuracy, our approach relies on a lightweight and simple architecture that consists of only standard neural networks layers. This makes our approach a natural fit for AR applications requiring real-time mobile inference. 

{
    \small
    \bibliographystyle{CVPR/ieeenat_fullname}
    \bibliography{main}
}

\newpage
\maketitlesupplementary

\appendix

In this appendix, we present additional results and ablations for our proposed MeshPose system. We also provide more technical details on our architecture and its training for reproducibility.

We begin in Section~\ref{sec:qualitative_eval} by showing more qualitative results across multiple image datasets (COCO~\cite{lin2014microsoft}, 3DPW~\cite{von2018recovering},  H36M~\cite{ionescu2013human3} and 3DOH~\cite{zhang2020object}) to demonstrate the wide applicability of our approach to multiple scenarios. We also provide more results on videos to showcase the temporal stability of our method even without temporal smoothing post-processing. We refer the readers to our \textit{mp4} video provided in the \textit{zip} file as our results are best viewed as videos.

Then, in Section~\ref{sec:sparsepose_ablations}, we ablate - with more metrics - the design choice for our novel VertexPose module that leverages DensePose~\cite{guler2018densepose} annotations to learn 2d vertex localization. We show first how our vertex-based representation (VertexPose) compares to the UV-based representation introduced in DensePose~\cite{guler2018densepose} in terms of DensePose metrics. Then, we evaluate multiple strategies to aggregate vertex UVs into pixel UVs to show the superiority of the barycentric UV aggregation.

In Section~\ref{sec:occlusion}, we further evaluate our system with respect to occlusion on the 3DOH and 3DPW-OCC datasets. We demonstrate that MeshPose is robust to occlusion and on par with other methods.

Then, we provide a more comprehensive 2d evaluation of our approach with other competing methods by reporting more metrics in Section~\ref{sec:2d_evaluation}.

In Section~\ref{sec:mobile}, we quantitatively demonstrate our real-time inference speed on mobile device making our approach a prime candidate for AR applications.

Finally, in Section~\ref{sec:architecture}, we provide further details on our architecture and its training. We detail the architecture of our backbone networks, our losses and our decoding strategy allowing us to predict high resolution meshes from the low-poly topology used throughout our pipeline.

\section{Qualitative Evaluation}
\label{sec:qualitative_eval}

\subsection{Visualizations on COCO}
\label{sec:coco_vis}
\begin{figure}
    \includegraphics[width=\linewidth]{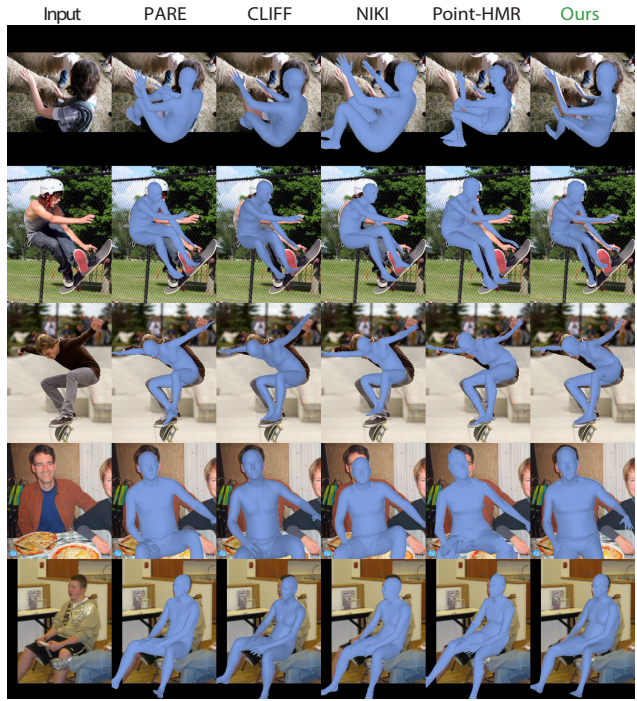} 
    \caption{Qualitative comparison on COCO against 4 state-of-the-art mesh reconstruction systems. MeshPose is consistently more aligned to the silhouette of the person.}
    \label{fig:coco_visualization}
\end{figure}

In Figure~\ref{fig:coco_visualization}, we showcase more results on the COCO~\cite{lin2014microsoft} dataset. Our approach demonstrates the ability to generate image-aligned meshes even in challenging scenarios such as occlusion or truncation of the body, which are common failure modes for other human mesh recovery systems. Unlike parametric methods bound to SMPL~\cite{loper2015smpl} models, our non-parametric mesh prediction approach, combined with DensePose supervision, offers greater flexibility and accurately captures very diverse body shapes that previous models struggle with.

\subsection{Visualizations on 3DPW}
\label{sec:3dpw_vis}
\begin{figure}
    \includegraphics[width=\linewidth]{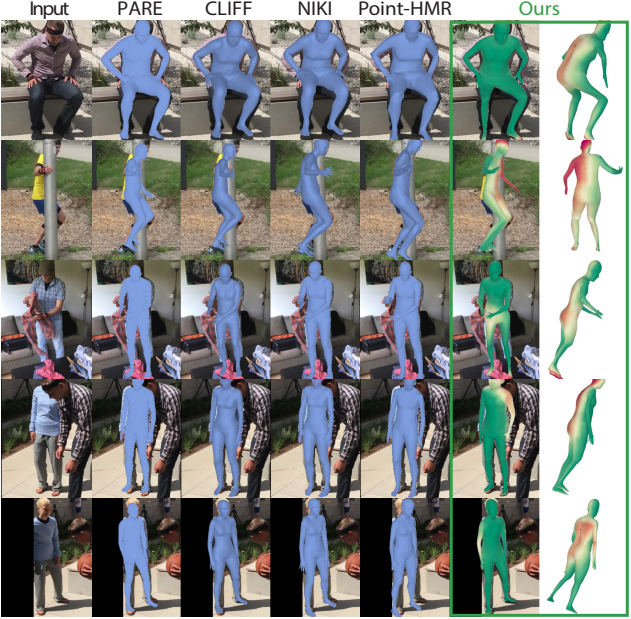} 
    \caption{Qualitative comparison on 3DPW (occluded subset) against 4 state-of-the-art mesh reconstruction systems. We also display the rendered visibility weights predicted by our system with a color map ranging from green (fully visible) to red (non visible).}
    \label{fig:3dpw_visualization}
\end{figure}

We present additional visualizations of our 3D mesh reconstruction on the 3DPW~\cite{von2018recovering} dataset in Figure~\ref{fig:3dpw_visualization}. In contrast to COCO~\cite{lin2014microsoft}, 3DPW showcases fewer occlusions, resulting in more full body meshes, and we thus focus our visualisations on 3DPW occluded subset. We show that our 3D mesh reconstruction is also competitive in these scenarios with more accurate 2D reprojection (see elbows, shoulders, limbs) while offering strong depth prediction. We also present the rendered visibility weights predicted by our system with a color map ranging from green (fully visible - predicted by the VertexPose branch) to red (non visible - predicted by the regression branch).

\subsection{Visualizations on H36M}
\label{sec:h36m_vis}
In Figure~\ref{fig:h36m_3doh_visualization} (top), we display our 3D mesh reconstruction performance on the H36M~\cite{ionescu2013human3} motion-capture dataset. We observe similar performance on this dataset that - similarly to 3DPW - exhibits only few occlusions.

\subsection{Visualizations on 3DOH}
\label{sec:3doh_vis}
\begin{figure*}
    \center
    \includegraphics[width=0.6\textwidth]{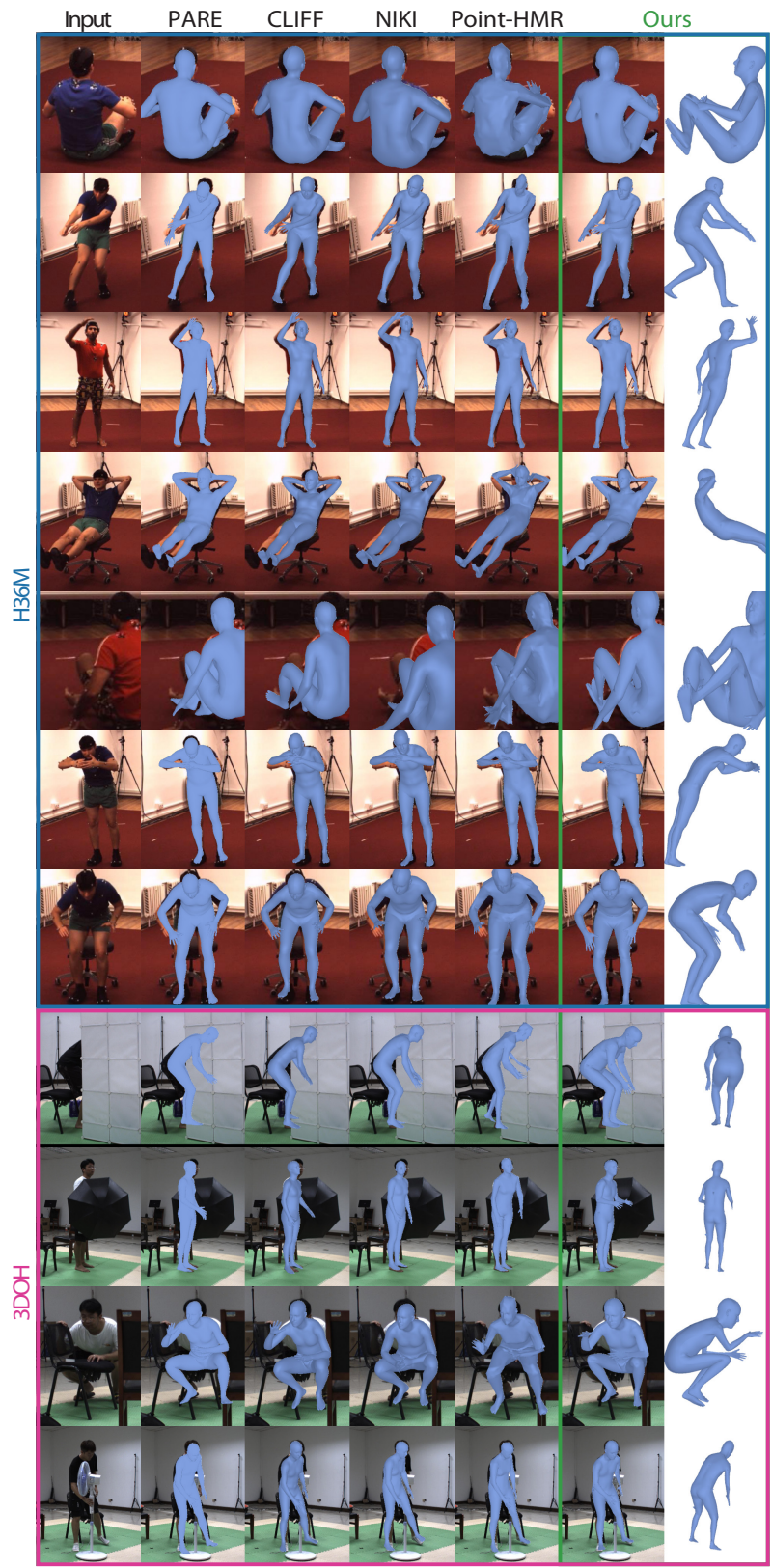}
    \caption{Qualitative comparison on H36M and 3DOH against 4 state-of-the-art mesh reconstruction systems. MeshPose exhibits stronger 2d reprojection accuracy especially on complex regions (elbows, shoulders, limbs).}
    \label{fig:h36m_3doh_visualization}
\end{figure*}

To demonstrate the robustness of our approach with respect to occlusion, we further showcase 3D mesh reconstruction visualisations on the 3DOH~\cite{zhang2020object} dataset in Figure~\ref{fig:h36m_3doh_visualization} (bottom). We compare our approach to other state-of-the-art methods in complex occluded scenes which are very common in real-world applications.

\subsection{Visualizations on Internet Videos}
\label{sec:pexel_vis}
\begin{figure}
    \includegraphics[width=\linewidth]{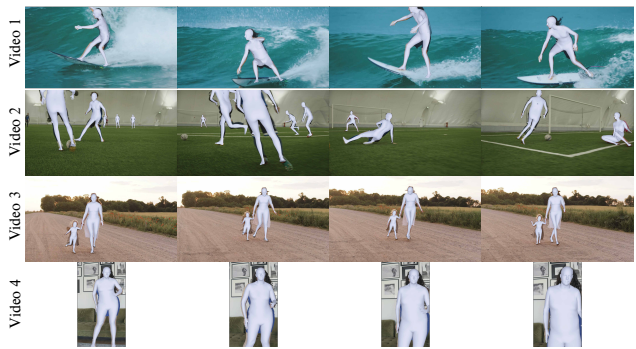} 
    \caption{Frames extracted from the videos available on our website. Results are best viewed on videos.}
    \label{fig:video_visualization}
\end{figure}

To provide a complete qualitative analysis, we also evaluate our approach on videos of humans in action \cite{pexels}. A concatenated video is available on our website. We also show a collection of frames in Figure~\ref{fig:video_visualization} but we recommend watching the results on the videos. We demonstrate very strong temporal stability (low jitter) even when applied frame-by-frame without any post-processing. In the videos attached, only the detected bounding box is temporally-smoothed. Our method is lightweight and simple with real-time inference, making it a prime candidate for AR applications.

\section{Quantitative Evaluation}

In this section, we present an additional quantitative analysis. To isolate and eliminate potential inaccuracies coming from the bounding box predictor, each network is assessed using ground truth bounding boxes.

\subsection{VertexPose Ablation Study}
\label{sec:sparsepose_ablations}

We provide more metrics for Table~2 introduced in the main paper. We report the standard Average Precision \textbf{AP} and Average Recall \textbf{AR}. We also provide $\textbf{AP}_{50}$ and $\textbf{AP}_{75}$ which measure precision at 50\% and 75\% IoU thresholds, $\textbf{AP}_M$ and $\textbf{AP}_L$ that evaluate precision for medium and large objects. The same metrics are also used for \textbf{AR}. First, in order to better understand the impact of training with vertex-based (VertexPose) representation compared to UV-based (DensePose~\cite{guler2018densepose}) representation, we compare both approaches in \reftab{table:dp_vs_sp_appendix} evaluated in terms of DensePose metrics. For a fair comparison, we re-trained DensePose with our backbone and our training settings. We closely follow the DensePose multi-head architecture introduced in \cite{guler2018densepose}. More specifically, for each pixel, we predict (i) the foreground segmentation mask $I$ via the classification branch and (ii) the patch label $c$ and the corresponding $[U,V]$ on that patch via the regression branch. The patch label $c^{*}$ is predicted by $P$ a 25-way (24 patches and background) classification unit $c^{*} = \arg\max_c P(c|i)$ while the UV is predicted by the UV regressor $R_{c^{*}}$ of the predicted patch $c^{*}$, $[U,V] = R_{c^{*}} (i)$. We observe that the VertexPose-based results compare favorably to their DensePose-based counterparts for all metrics, thus confirming the merit of the proposed approach. The same results are observed across all tested architecture (MobileNet~\cite{mbnet}, ResNet~\cite{he2016deep}, HRNet32 and HRNet48~\cite{wang2020deep}).

In \reftab{table:decoding_appendix} we analyze the impact of the barycentric interpolation strategy proposed in Section~3.1.1 of the main paper. As a reminder, to decode the pixel UV, we compute the barycentric combination of the UV values of the vertices belonging to the face with the largest score $f^{*} = \arg\max_{f \in \mathcal{F}} \sum_{n \in f} q_n$. Here $q_v$ corresponds to the per-pixel posterior over vertices:
\begin{align}
q_v = \frac{\exp(\sparseposepix_v)}{\sum_{k=1}^{V}{\exp(\sparseposepix_k)}}
\end{align}

We thus compare our proposed approach to simpler baselines: (i) decoding the pixel UV as the UV of the strongest vertex at any given pixel (`Nearest') and (ii) decoding the UV via argsoftmax over all vertices rather than those of the strongest triangle (`Global Average'):
\begin{align}
\uv = \sum_{m \in \mathcal{V}} \beta_m \uv_m, ~\mathrm{where}~ \beta_m = \frac{q_m}{\sum_{n \in \mathcal{V}} q_n}
\end{align}
with $\mathcal{V}$ the set of all vertices. The evaluation indicates the merit of the smooth transition between vertices secured by barycentric interpolation on almost all DensePose metrics.

\begin{table*}[!htp]\centering
\begin{subtable}[c]{\textwidth}
\centering
\begin{tabular}{c|ccccc|ccccc}
  \toprule
    & $\textbf{AP}$ & $\textbf{AP}_{50}$ & $\textbf{AP}_{75}$ & $\textbf{AP}_{M}$ & $\textbf{AP}_{L}$ & $\textbf{AR}$ & $\textbf{AR}_{50}$ & $\textbf{AR}_{75}$ & $\textbf{AR}_{M}$ & $\textbf{AR}_{L}$\\
   \midrule
DP - MBNet & 53.54 & 91.66 & 58.29 & 54.75 & 53.65 & 64.09 & 95.68 & 75.39 & 54.89 & 64.73 \\
SP - MBNet & 54.08 & 93.01 & 59.41 & 55.94 & 54.32 & 64.46 & 96.03 & 76.10 & 56.24 & 65.03 \\
\midrule
DP - ResNet & 57.31 & 93.06 & 65.60 & 57.72 & 57.49 & 67.51 & 96.30 & 80.29 & 57.87 & 68.17 \\
SP - ResNet & 58.87 & 93.05 & 68.34 & 60.10 & 58.90 & 68.73 & 96.26 & 82.08 & 60.14 & 69.33 \\
\midrule
DP - HRNet32 & 61.12 & 94.84 & 72.23 & 61.39 & 61.41 & 70.53 & 97.33 & 84.26 & 61.56 & 71.16 \\
SP - HRNet32 & 61.24 & 94.63 & 72.61 & 61.68 & 61.54 & 70.52 & 97.10 & 84.53 & 61.77 & 71.13 \\
\midrule
DP - HRNet48 & 62.74 & 95.04 & 75.39 & 63.86 & 63.03 & 71.95 & 97.50 & 86.18 & 64.11 & 72.49 \\
SP - HRNet48 & \textbf{63.32} & \textbf{95.12} & \textbf{76.48} & \textbf{64.75} & \textbf{63.63} & \textbf{72.14} & \textbf{97.73} & \textbf{87.07} & \textbf{64.89} & \textbf{72.65} \\
\bottomrule
\end{tabular}
\caption{VertexPose (VP) vs DensePose (DP).}\label{table:dp_vs_sp_appendix}
\end{subtable}
\hfill
\begin{subtable}[c]{\textwidth}
\centering
 \begin{tabular}{cccccccccccc}
  \toprule
   & $\textbf{AP}$ & $\textbf{AP}_{50}$ & $\textbf{AP}_{75}$ & $\textbf{AP}_{M}$ & $\textbf{AP}_{L}$ & $\textbf{AR}$ &
   $\textbf{AR}_{50}$ & $\textbf{AR}_{75}$ & $\textbf{AR}_{M}$ &
   $\textbf{AR}_{L}$\\
   \midrule
Barycentric & \textbf{61.24} & 94.63 & \textbf{72.61} & \textbf{61.68} & \textbf{61.54} & \textbf{70.52} & 97.10 & \textbf{84.53} & \textbf{61.77} & \textbf{71.13} \\
Closest & 59.69 & \textbf{95.10} & 70.08 & 61.39 & 60.01 & 68.80 & \textbf{97.59} & 83.24 & 61.49 & 69.31 \\
Global Average & 33.35 & 89.58 & 11.00 & 37.13 & 33.61 & 43.23 & 94.38 & 31.79 & 37.16 & 43.64 \\
\bottomrule
\end{tabular}
\caption{UV aggregation strategy}\label{table:decoding_appendix}
\end{subtable}
\caption{
Analysis of VertexPose performance on the DensePose-COCO dataset. We evaluate the impact of the backbone choice and the UV decoding strategy.
}
\end{table*}

\subsection{Occlusion Evaluation}
\label{sec:occlusion}
\begin{table}
\centering
\resizebox{\columnwidth}{!}{
    \begin{tabular}{l|r|r|r|r|r}
        \toprule
        \bf Method & \multicolumn{3}{c}{\bf 3DPW-OCC $\downarrow$} & \multicolumn{2}{|c}{\bf 3DOH $\downarrow$} \\
        \cmidrule(lr){2-6}
        & MPJPE & PAMPJPE & PVE & MPJPE & PAMPJPE \\ 
        \midrule        
        DOH~\cite{zhang2020object} & - & 72.2 & - & - & 58.5\\
        CRMH~\cite{jiang2020mpshape} & - & 78.9 & - & - & -\\
        VIBE~\cite{kocabas2020vibe} & - & 65.9 & - & - & -\\
        SPIN~\cite{kolotouros2019learning} & 95.6 & 60.8 & 121.6 & 104.3 & 68.3\\
        HMR-EFT~\cite{joo2020eft} & 94.4 & 60.9 & 111.3 & 75.2 & 53.1\\
        HybrIK~\cite{li2021hybrik} & 90.8 & 58.8 & 111.9 & 40.4 & 31.2\\
        PARE~\cite{kocabas2021pare} & 90.5 & 56.6 & \textbf{107.9} & 63.3 & 44.3\\
        ROMP~\cite{sun2021monocular} & - & 67.1 & - & - & - \\
        NIKI~\cite{li2023niki} & 88.2 & 55.3 & 109.7 & \textbf{38.9} & \textbf{29.2} \\
        PLIKS~\cite{pliks} & \textbf{86.1} & 53.2 & - & 51.5 & 39.3 \\
        \midrule
        MeshPose (HRNet32) & 89.11 & \textbf{51.81} & 108.15 & 60.93 & 38.18\\    
        \bottomrule
    \end{tabular}
}
    \caption{Evaluation of the occlusion robustness of MeshPose compared to other state-of-the-art approaches in object-occluded benchmark datasets 3DPW-OCC and 3DOH.} 
    \label{tab:occlusion_table}
\end{table}

To further validate the stability of our proposed method under occlusion, we also conducted evaluation on the object-occluded benchmark dataset: 3DPW-OCC\cite{von2018recovering, zhang2020object} and 3DOH50K\cite{zhang2020object}. \textbf{3DPW-OCC} refers to a new test-set with occluded sequences from the entire 3DPW dataset, while \textbf{3DOH} is a 3D human dataset with human activities occluded by objects, which  provides 2D, 3D annotations and SMPL parameters.
For a fair comparison following previous methods we train our model by including 3DOH without 3DPW (since some videos of 3DPW-OCC are from the training set).

From the evaluation results shown in  Table~\ref{tab:occlusion_table}, we see that the proposed MeshPose performs also very well in occluded scenarios by achieving state-of-the-art performance and outperforming approaches that are designed to deal with occlusion. Moreover, in Figure~\ref{fig:h36m_3doh_visualization}, we show that MeshPose achieves good mesh reconstructions even in cases with heavy object-occlusions. These results demonstrate the effectiveness of our proposed method that is able to deliver pixel-aligned 3D-mesh reconstructions and at the same time deal successfully with occlusion.

\subsection{2D Evaluation Benchmark}
\label{sec:2d_evaluation}

In Table~\ref{tab:2d_baseline_deepdown}, we expand the analysis from the main paper with additional evaluations with 2D metrics. For sake of completeness, we include LVD~\cite{corona2022learned}, though this method is not directly comparable as it has not been trained with in-the-wild datasets. We present the Average Precision (\textbf{AP}) and Average Recall (\textbf{AR}) for 2D COCO keypoints in instances where at least 80\% of the keypoints are visible. This 80\% threshold delineates a sub-dataset with instances that are almost fully visible, but yet remains sufficiently large to allow for meaningful conclusions. $\textbf{AP}_M$ and $\textbf{AP}_L$ measure medium (area between $32^2$ and $96^2$ pixels in the image) scale and large scale (area above $96^2$ pixels in the image) object detection precision, while $\textbf{AR}_M$ and $\textbf{AR}_L$ assess recall for medium and large scale objects. Additionally, we provide the \textbf{AP} and \textbf{AR} metrics for the DensePose task.

Our approach surpasses the other methods in terms of 2D metrics, showcasing its strong 2D alignment.

\begin{table*}
\centering

    \begin{tabular}{ll|c|c|c|c|c|c|c|c}
        \toprule
        & \bf Method & \multicolumn{6}{|c}{\bf 2D COCO KeyPoints $\uparrow$ 80\%} & \multicolumn{2}{|c}{\bf DensePose $\uparrow$} \\
        \cmidrule(lr){3-10}
        &  & $\textbf{AP}$ & $\textbf{AP}_M$ & $\textbf{AP}_L$ & $\textbf{AR}$ & $\textbf{AR}_M$ & $\textbf{AR}_L$ & $\textbf{AP}$ & $\textbf{AR}$ \\ 
        \midrule
        \parbox[t]{2mm}{\multirow{6}{*}{\rotatebox[origin=c]{90}{param}}} 
            & DaNet~\cite{zhang2020densepose2smpl} & 52.60 & 51.00 & 54.10 & 65.20 & 62.80 & 66.50 & 16.42 & 29.24\\
            & HybrIK~\cite{li2021hybrik} & 31.70 & 25.40 & 35.30 & 47.50 & 37.10 & 52.80 & 20.85 & 34.44\\
            & PARE~\cite{kocabas2021pare} & 56.90 & 56.30 & 57.70 & 67.20 & 64.80 & 68.50 & 30.02 & 41.54\\
            & CLIFF~\cite{li2022cliff} & 61.40 & 58.80 & 63.70 & 72.50 & 68.10 & 74.70 & 30.99 & 41.83\\
            & PyMaf~\cite{pymaf2021} & 58.00 & 57.70 & 58.70 & 70.00 & 67.50 & 71.30 & 17.62 & 30.69\\
            & LVD~\cite{corona2022learned} & 12.20 & 6.60 & 13.20 & 28.90 & 18.60 & 29.90 & 1.03 & 6.49\\
            & NIKI~\cite{li2023niki} & 48.50 & 44.80 & 51.10 & 61.50 & 55.70 & 64.50 & 25.11 & 37.39\\
        \midrule
        \parbox[t]{2mm}{\multirow{4}{*}{\rotatebox[origin=c]{90}{n-param}}}
            & Metro~\cite{lin2021end} & 30.40 & 33.00 & 29.60 & 45.10 & 45.00 & 45.10 & 9.28 & 21.16\\
            & Graphormer~\cite{lin2021mesh} & 35.90 & 37.00 & 35.70 & 49.50 & 48.20 & 50.00 & 13.57 & 26.20\\
            & FastMetro~\cite{cho2022FastMETRO} & 39.30 & 40.60 & 39.30 & 53.20 & 52.00 & 53.80 & 13.76 & 26.21\\
            & PointHMR~\cite{PointHMR} & 44.40 & 42.80 & 46.10 & 57.30 & 53.10 & 59.40 & 19.58 & 32.18\\
        \midrule
        \parbox[t]{2mm}{\multirow{3}{*}{\rotatebox[origin=c]{90}{ours}}}
            & MeshPose (HRNet32 \cite{wang2020deep}) & \textbf{71.20} & \textbf{67.20} & \textbf{73.70} & \textbf{79.60} & \textbf{74.00} & \textbf{82.50} & \textbf{47.87} & \textbf{57.62}\\    
            & MeshPoseS (ResNet50 \cite{he2016deep}) & 67.00 & 63.20 & 69.50 & 76.50 & 70.80 & 79.40 & 44.41 & 54.49\\
            & MeshPoseXS (MBNet140 \cite{mbnet}) & 63.60 & 59.00 & 66.50 & 73.90 & 67.90 & 76.90 & 38.56 & 49.20\\
        \bottomrule
    \end{tabular}

\caption{Evaluation of 2D accuracy in COCO-DensePose for both 2D keypoint predictions and DensePose regression.}
    \label{tab:2d_baseline_deepdown}
\end{table*}

\subsection{Inference setup and mobile inference}
\label{sec:mobile}

For our desktop inference experiment, we assess the performance speed of the backbone network of each baseline methods using the original authors' official implementations. Each timing was evaluated on the same machine equipped with a Nvidia Tesla V100 GPU.

Our models are purely convolutional and thus run out-of-the-box on modern phones with accelerators. We exported the ONNX \cite{onnx} versions of our models and computed their timings (FPS) on an iPhone-12 using the CoreML \cite{coreml} backend, obtaining comparable timings to the GPU-desktop timings: $97$, $99$, and $153$ FPS for Meshpose, MeshposeS, and MeshposeXS respectively.

\section{Architecture and training details}
\label{sec:architecture}
In order to provide a comprehensive understanding of our method outlined in the main paper, we present additional details regarding the design of our pipeline and the training process. We begin by providing more details on the design of our backbones in Subsection~\ref{sec:backbone}. Following that, we include supplementary details concerning our multi-head decoders in Subsection~\ref{sec:decoders}. More precisely, we detail our vertex and visibility regression branch (\ref{sec:regression}), our custom silhouette rendering - both introduced in Section~3.2.1 of the main paper (\ref{sec:renderingsil}) and our high-poly mesh upsampler presented in Section~3.2.3 of the main paper (\ref{sec:upsampler}). Finally, in Subsection~\ref{sec:training} we provide more details on our training strategy including datasets mixing, augmentations and scheduling.

\subsection{Backbone architectures}
\label{sec:backbone}
Regarding our main backbone architecture of choice we employed the HRNet-32 model described in \cite{wang2020deep}, as it is capable of producing a high-resolution feature map. Only the high-resolution, stride $4$ output features of the last block are used. Since MeshPose is a multitasking system which outputs tensors of a large dimensionality, we have modified the architecture to have an output with more feature channels, without adding considerable overhead. More specifically, before the upsampling and the sum-based fusion of the feature maps from the last block, which have different stride and feature size, we project all of them to a fixed feature size of $256$ instead of $32$ that is used in the original HRNet-32 implementation. This modification only adds a small number of parameters, since it is applied only at the end of the backbone, but it removes the $32$-channel bottleneck to accomodate the MeshPose tasks. 

For the Resnet50 \cite{he2016deep} and MobileNetV2 \cite{mbnet} variants we used dilated convolutions on their last block, which gives features with stride $16$, and then we applied a decoder-net with separable-convolutions and skip-connections from the previous stages in order to produce the final feature map with stride $4$. We found that this light-weight decoder-net has much less parameters (due to separable convolutions) compared to the fully deconvolutional layers that are usually employed in pose estimation \cite{xiao2018simple}.

\subsection{Decoders}
\label{sec:decoders}
As explained in Section~3.2 of the main paper, we learn how to directly regress 3D vertex positions $V_{reg}^{XYZ}$ for all vertices. In addition to the 3D positions, we also predict a visibility label $w\in [0,1]$ for each vertex. The visibility $w$ indicates whether we should rely on the VertexPose-based 2D position, $V_{sp}^{XY}$ or fall back to the $V_{reg}^{XY}$ value. A low $w$ value implies that the corresponding vertex is occluded or out-of-crop which means that the VertexPose vertex is likely incorrect. The final MeshPose vertices are simply computed as a visibility-weighted average between both predictions: $V_{mp}^{XY} =  V_{sp}^{XY}  w + V_{reg}^{XY} (1-w)$. In this subsection, we detail how the regressed vertices $V_{reg}^{XYZ}$ are obtained and how we supervise them together with their visibility predictions $w$.

\subsubsection{Coordinate and Visibility Regression}
\label{sec:regression}
We draw inspiration from \cite{pliks} and extract the last tensor $\mathbf{F}\in\mathbb{R}^{C,\frac{H}{4},\frac{W}{4}}$ of our CNN backbone with $C=256$. As explained in the main paper, we use 1D convolutions to generate three feature tensors $\mathbf{P}=\{\mathbf{P}^x,\mathbf{P}^y,\mathbf{P}^z\}\in\mathbb{R}^{V\times 64}$, one for each dimension $X$, $Y$, $Z$ and an extra visiblity weight $\mathbf{w}\in\mathbb{R}^{V}$, with $V$ the number of vertices.

\begin{align}
\mathbf{P}^x &= f_x^{1D}(\psi_x(\text{avg}^{x,y}(\mathbf{F})))\\
\mathbf{P}^y &= f_y^{1D}(\psi_y(\text{avg}^{x,y}(\mathbf{F})))\\
\mathbf{P}^z &= f_z^{1D}(\psi_z(\text{avg}^{x,y}(\mathbf{F})))
\end{align}
and
\begin{align}
\mathbf{w} &= \sigma(\text{avg}^z(f_z^{1D}(\psi_z(\text{avg}^{x,y}(\mathbf{F})))))
\end{align}

We first apply an average pooling $\text{avg}^{x,y}$ across the spatial dimensions $(x,y)$. Then, we apply two successive 1D convolutions $\psi_{i}$ and $f_{i}^{1D}$ along the indexed dimension $i$. The first convolution $\psi_{i}$ expands the dimension from $C\times 1$ to $C\times C^\prime$ then $f_{i}^{1D}$ transforms the feature tensors to $V\times C^\prime$ dimension. Finally, for the visibility weight $\mathbf{w}$, we average across the channel dimension $\text{avg}^z$, then apply a sigmoid $\sigma$ activation function to map the values between 0 and 1.

To obtain the 3D vertex positions from the learnt features $\{\mathbf{P}^x,\mathbf{P}^y,\mathbf{P}^z\}$, we apply argsoftmax over the $C^\prime=64$ channels. The resulting value of the argsoftmax will thus be between $0$ and $64$ and thus needs to be mapped to pixel positions. We map the range $[0,64]$ to $[-W,2W]$ for $X$, $[-H,3H]$ for $Y$ and $[-2W,2W]$ for $Z$. The top left pixel on the image corresponds to pixel $[0,0]$. The new range expands beyond the image boundary to predict out-of-crop vertices. We note that to accommodate for selfie-like images, we consider a larger range for $Y$ ($[-H,3H]$): this allows us to be able to predict the position of leg vertices that will often lie significantly below the crop.

\subsubsection{Custom Silhouette Rendering}
\label{sec:renderingsil}
\begin{figure}
    \centering
\subfloat[\centering Edge and Vertex Loss]{{
 \includegraphics[width=.21\textwidth]{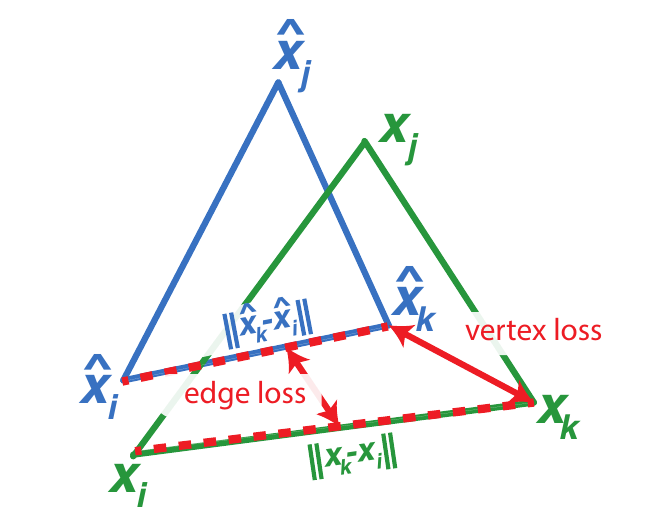}}}
  \qquad
   \subfloat[\centering Joint Localization Loss]{{ \includegraphics[width=.21\textwidth]{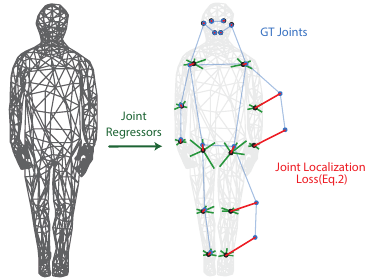}}}
    \caption{
     Regularization losses used for 3D mesh supervision.}
    \label{fig:3d_losses}
\end{figure}

The learnt visibility weight $\mathbf{w}$ is partly supervised by the 3D localization and the edge losses (see Section~3.2.4 and Figure~\ref{fig:3d_losses}). We also use a binary-cross entropy loss $L_{W}$ using the supervision of the mesh pseudo-ground truth. However, we also want to leverage the ground truth DensePose segmentation masks which provide a suitable signal to learn visibility with weak supervision. To achieve that, we introduce a novel silhouette rendering module by modifying the soft rasterization method of SoftRas~\cite{liu2019softras} so that it incorporates the predicted vertex visibilities. More specifically, for each pixel $i$, we compute the silhouette $I_s$ by:
\begin{equation}
I_s^i = \mathcal{A}_O(\{\mathcal{D}_j\}) = 1- \prod_j (1 - w_j^i\mathcal{D}_j^i)
\end{equation}
Here, as in~\cite{liu2019softras}, $\mathcal{D}_j$ denotes the influence of triangle $f_j$ at pixel $i$ and mostly depends on the distance of triangle $j$ to pixel $i$. Contrary to~\cite{liu2019softras}, we also multiply the influence $\mathcal{D}_j$ by the visibility weight $w_j^i$ of face $j$ at pixel $i$. $w_j^i$ is simply computed as the linear interpolation between the visibility weights of the three vertices of face $j$ at pixel $i$. The newly added coefficient $w_j^i$ modifies the initial rendering pipeline from~\cite{liu2019softras} to ignore faces with low visibility weight.

We use two losses to supervise our rendered silhouette $I_s$. First, we use a simple $L_2$ loss between our rendered $I_s$ and the corresponding ground truth from the DensePose annotation $I_s^{DP}$: $\mathcal{L}_{I_s}^1=||I_s-I_s^{DP}||_2$.
The second loss we introduce is inspired from the boundary loss from~\cite{pmlr-v102-kervadec19a}:
\begin{align}
\mathcal{L}_{I_s}^2=-\sum_{x,y }D(x,y) I_s(x,y)
\end{align}
with $D$ the level set of the DensePose ground-truth segmentation boundary. More specifically, $D(x,y)$ is equal to the distance $d$ between pixel $(x,y)$ and the boundary $\partial I_s^{DP}$ of the ground-truth segmentation mask - except for pixels $(x,y)$ outside of the ground truth mask.
\begin{align}
D(x,y)\left\{
    \begin{array}{ll}
        d((x,y), \partial I_s^{DP}) & \mbox{if } I_s^{DP}(x,y)=1\\
        0 & \mbox{else.}
    \end{array}
\right.
\end{align}
This loss only penalizes "too small" silhouettes that are not expanding over the whole ground truth mask. The final silhouette loss is the combination of the two introduced losses:
\begin{align}
\mathcal{L}_{I_s}=100\cdot \mathcal{L}_{I_s}^1+\mathcal{L}_{I_s}^2
\end{align}

To also learn the visibility weight $\mathbf{w}$ for out-of-crop vertices, we render the silhouette on a larger crop and pad the ground truth mask accordingly. This helps the pipeline to assign low visibility weights to vertices that are not in the original crop.

\subsubsection{High Poly Mesh Upsampler}
\label{sec:upsampler}

As mentioned in the main paper, we predict the vertices of a low-polygon approximation ($518$ vertices) of our high resolution body mesh ($6890$ vertices). As pointed out in~\cite{lin2021end}, working on a lower-resolution mesh both reduces memory usage while improving training stability by limiting correlated vertices. To generate this approximation, we used a custom-defined mesh to reduce the number of vertices in undesirable high curvature areas (hands, fingers).

To obtain the high-poly mesh  from the low-poly mesh , which is the output of the network, we trained a multilayer perceptron upsampler following the ideas of~\cite{lin2021end, kolotouros2019convolutional}. We employed two affine layers to progressively upsample the low-poly mesh to the high-poly variant (with an intermediate representation of $1723$ vertices as in~\cite{lin2021end}). We applied an L1 loss between the ground-truth and the upsampled vertices at each stage, as well as between the GT joints and the joints that are estimated using a landmarker on the upsampled version. We trained the upsampler independently from the main network using the high-poly mesh pseudo GT annotations \cite{Moon_2022_CVPRW_NeuralAnnot} from only COCO, Human 3.6M and MPI-INF-3DHP datasets to accurately cover a large pose distribution. Our method is thus agnostic to the high poly template and only the upsampler would need to be tuned for a different template to be used.

An additional improvement to the training of the upsampler, compared to previous approaches, is the addition of synthetic noise during the training to make the upsampler more robust to noise or to small errors that may be in the low-poly mesh. More specifically, with probability $0.5$ we randomly added to $25\%$ of the vertices spikes equal to $15\%$ of their vertex position values (after applying root alignment to the mesh). In the other case, with probability $0.5$ we added gaussian noise with standard deviation $5\%$ to each vertex position.

For comparison, a nearest-point-on-triangle, non-learned upsampler produces slightly worse results (1mm difference for 3DPW MVE, 2 for DP AP).

\subsection{Training}
\label{sec:training}

We used a combination of datasets and training signals in our model training. More specifically, we used a mixture of COCO and Human Mesh Reconstruction (HMR) datasets. First, we enriched the COCO (train2017) dataset with mesh pseudo-annotations and their DensePose annotations. For the HMR datasets, we combined Human3.6M~\cite{ionescu2013human3}, MPI-INF-3DHP~\cite{mehta2017monocular} and 3DPW~\cite{von2018recovering} training sets. For each dataset, we proceeded to image augmentation with (i) flipping , (ii) rotation (between $-45^{\circ}$ and $45^{\circ}$), (iii) scale (between $0.75$ and $1.25$) and (iv) color augmentations. We followed a $\sim40/60$ split between COCO and HMR datasets. To achieve that, we upsampled each dataset by an associated factor to control the dataset mixture: $5$ for COCO, $4$ for Human3.6M, $1$ for MPI-INF-3DHP and $2$ for 3DPW. We trained with the Adam optimizer for $200$ epochs with a mini-batch size of $96$ and a learning rate of $1 \times 10^{-3}$, which is reduced by a factor of $10$ after $100$ epochs. The weights of the backbone of our systems are initialized with pretrained 2D pose estimation networks. We used Linux machines with 4 Nvidia Tesla V100 GPUs (16GB) for all of our experiments.

We aggregated the different losses based on the following weighted linear combination:
\begin{align}
\mathcal{L} = \mathcal{L}_{BL} + L_{\mathrm{consistency}} +10\cdot L_W+0.1\cdot\mathcal{L}_{V}\nonumber\\+\mathcal{L}_{E}+0.1\cdot\mathcal{L}_{N}+\mathcal{L}_{J}+ \mathcal{L}_{I_s}
\end{align}
with $\mathcal{L}_{BL}$ the barycentric cross-entropy loss (Section~3.1.2), $L_{\mathrm{consistency}}$ the UV consistency loss (Section~3.1.2), $L_W$ the visibility binary cross-entropy loss (Section~\ref{sec:renderingsil}), $(\mathcal{L}_{V},\mathcal{L}_{E},\mathcal{L}_{N})$ the vertex localization, the edge, the normal and the joint localization losses (Section~3.2.4) and $\mathcal{L}_{I_s}$ the silhouette loss (Section~\ref{sec:renderingsil}).

We note that $(\mathcal{L}_{V},\mathcal{L}_{N})$ are given smaller weights to downscale the importance of the pseudo-ground truth. Our system however requires higher weight for $\mathcal{L}_{E}$ to remove mesh artefacts.

\end{document}